%% file: acl_latex.tex
\pdfoutput=1

\documentclass[11pt]{article}

\usepackage[preprint]{acl}
\usepackage{amsmath}
\usepackage{amssymb}
\usepackage{times}
\usepackage{latexsym}
\usepackage{subcaption}
\usepackage{amsthm}

\theoremstyle{definition}

\usepackage[T1]{fontenc}

\usepackage[utf8]{inputenc}
\usepackage{wrapfig}

\usepackage{microtype}

\usepackage{inconsolata}
\usepackage[most]{tcolorbox}
\tcbset{
  myhighlight/.style={
    colback=yellow!5!white,
    colframe=yellow!50!black,
    boxrule=0.5pt,
    arc=3pt,
    left=6pt,
    right=6pt,
    top=4pt,
    bottom=4pt,
    fonttitle=\bfseries,
    coltitle=black
  }
}

\usepackage{graphicx}

\usepackage{algorithm}
\usepackage{algpseudocode}

\title{Emergence of Primacy and Recency Effect in Mamba: \\ A Mechanistic Point of View}

\author{
\textbf{Muhammad Cendekia Airlangga$^{*}{1}$, Hilal AlQuabeh$^{*}{1,2}$}\\
\textbf{Munachiso Samuel Nwadike$^{1,2}$, Kentaro Inui$^{1,2,3}$} \\
$^{1}$MBZUAI \quad $^{2}$RIKEN \quad $^{3}$Tohoku University\\ 
\texttt{\footnotesize muhammad.airlangga@mbzuai.ac.ae} 
}
\begin{document}

\maketitle
\begin{abstract}
We study memory in state-space language models using primacy and recency effects as behavioral tools to uncover how information is retained and forgotten over time. Applying structured recall tasks to the Mamba architecture, we observe a consistent U-shaped accuracy profile, indicating strong performance at the beginning and end of input sequences. We identify three mechanisms that give rise to this pattern. First, long-term memory is supported by a sparse subset of channels within the model's selective state space block, which persistently encode early input tokens and are causally linked to primacy effects. Second, short-term memory is governed by delta-modulated recurrence: recent inputs receive more weight due to exponential decay, but this recency advantage collapses when distractor items are introduced, revealing a clear limit to memory depth. Third, we find that memory allocation is dynamically modulated by semantic regularity: repeated relations in the input sequence shift the delta gating behavior, increasing the tendency to forget intermediate items. We validate these findings via targeted ablations and input perturbations on two large-scale Mamba-based language models: one with 1.4B and another with 7B parameters.

\end{abstract}

\input{sections/introduction.tex}

\input{sections/related_work}

\input{sections/task}

\input{sections/results}

\input{sections/conclusions}

\section*{Limitations}
Our study focuses on memory behavior in Mamba using primacy and recency effects within a controlled serial recall task. Although this setup provides clear interpretability, it abstracts away from the complexity of natural language understanding and may not fully capture how memory operates in real-world tasks such as question answering, summarization, or reasoning. 
These limitations suggest promising directions for extending our framework to broader settings and validating its relevance in downstream applications.

\section*{Acknowledgments}
This work was written independently, with minor phrasing assistance from a large language model (ChatGPT). 

\bibliographystyle{acl_natbib}

\input{acl_latex.bbl}
\appendix
\input{sections/appendix}

\end{document}

%% file: sections/introduction.tex
\section{Introduction}
\renewcommand{\thefootnote}{\fnsymbol{footnote}}
\footnotetext[1]{Equal contributions}
\footnotetext[2]{Code will be available upon publication at: \url{https://github.com/mcairlangga2/MambaInterp}}
\renewcommand{\thefootnote}{\arabic{footnote}}  
\label{sec:introduction}

Memory effects like primacy and recency—the improved recall of items at the beginning and end of a sequence—are central to cognitive psychology \citep{ebbinghaus1885memory, atkinson1968human}. Primacy reflects deeper encoding of early inputs, while recency arises from their availability in short-term memory.

Interestingly, large language models (LLMs) exhibit similar biases. In transformer-based models, primacy and recency emerge from architectural features such as causal masking and relative positional embeddings \citep{janik2023aspects, wu2025emergence}. These lead to phenomena such as the attention sink, where attention concentrates on early tokens \citep{gu2025attentionsinkemergeslanguage}, while recency arises from diluted attention and induction head failures.

\begin{figure*}[t]
  \centering
  \includegraphics[width=2\columnwidth]{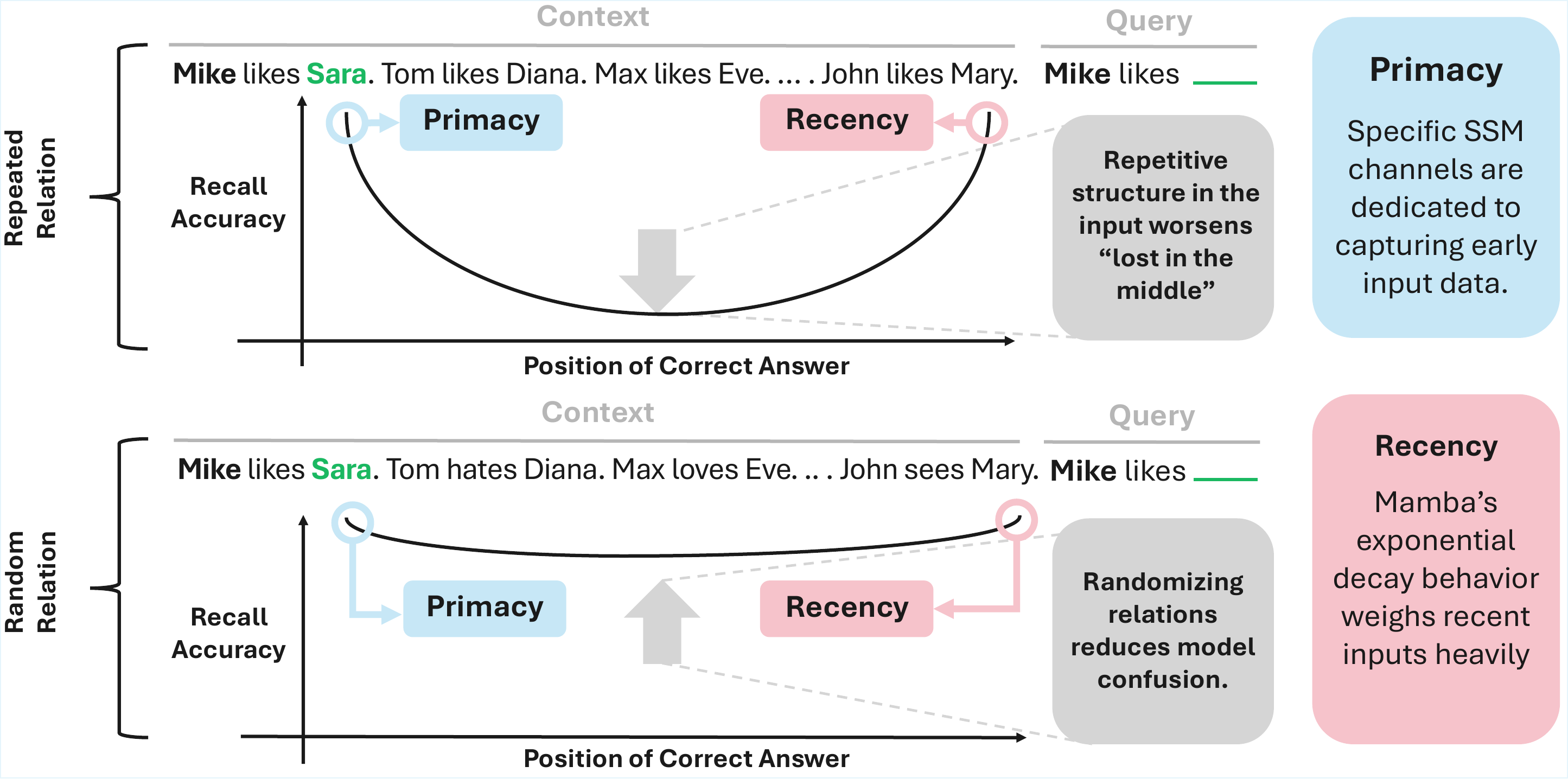}
  
  \caption{
\textbf{Illustration of the structured recall task with two variants: repeated relation (top) and random relation (bottom).}  
The task measures Mamba’s recall accuracy across sequence positions.. \textbf{Repeated relation} reuses the same predicate (\textit{likes}), producing strong \textbf{primacy} and \textbf{recency} effects, with a drop in middle recall. Primacy arises from early-tuned SSM channels; recency from Mamba’s decay favors recent tokens. Repetition increases positional confusion, worsening the ``lost-in-the-middle'' effect. \textbf{Random relation} breaks this pattern, easing ambiguity and improving middle recall, though primacy and recency remain.
}

  \label{fig:recall-task}
\end{figure*}

Recent work shows that state-space models (SSMs), which use recurrent updates instead of attention, can also display these effects. In Mamba, exponential decay in hidden state updates emphasizes recent tokens, producing recency \citep{wang2025understandingmitigatingbottlenecksstate}, while primacy appears in S4 models despite no explicit attention \citep{gu2021efficiently, morita2025emergenceprimacyeffectstructured}.

However, while these observations confirm the existence of primacy and recency effects in different SSMs, they leave several important questions open. First, prior studies tend to examine these biases in isolation, either primacy or recency, but rarely their joint emergence within a single model. Second, these works often focus on empirical detection without offering causal or mechanistic explanations for why and how these effects arise. 

In this work, we study short- and long-term memory in state space language models using structured recall tasks (Figure~\ref{fig:recall-task}). Rather than merely confirming the primacy and recency effects, we analyze their underlying mechanisms. Although Mamba’s architecture naturally supports recency, we show it struggles with repeated patterns, leading to rapid forgetting. In contrast, we causally identify the internal components that encode long-term memory and show that disrupting them selectively impairs the recall of early inputs while preserving short-term memory.

\begin{itemize}
\item We provide causal evidence that the primacy in Mamba is supported by a sparse subset of internal channels in the selective state-space block. These channels persistently encode early tokens and function as a long-term memory store. Disrupting these channels significantly diminishes primacy effects (section \ref{sec:primacy}, Figure \ref{fig:ltm+intervention}).

\item We show that recency arises naturally from Mamba’s architecture due to the exponential decay behavior, which weights recent inputs more heavily. Further, we show that by adding distraction, the recent recall accuracy reduces significantly, mimicking short memory performance (section \ref{sec:recency}, Figure \ref{fig:recency}).

\item We observe a Ranschburg-like \citep{jahnke1969ranschburg} effect in Mamba, where repeated patterns increase forgetting. We attribute this to its discretization mechanism (Section~\ref{sec:semantic}, Figure~\ref{fig:kernels}).

\end{itemize}

%% file: sections/related_work.tex
\section{Related Work}

\subsection{Memory and Recall Biases in Transformers}

Transformer-based language models have been shown to exhibit memory phenomena such as primacy and recency effects when evaluated on structured recall tasks~\citep{janik2023aspects}. These effects emerge due to architectural asymmetries introduced by causal masking, learned positional embeddings, and attention dynamics~\citep{wu2025emergence}. A key mechanism behind long-range recall in transformers is the \textit{induction head}, which allows the model to learn copy operations by attending to repeated patterns in the input~\citep{olsson2022context}. 


Primacy in transformers has been linked to the \textit{attention sink}, where early tokens attract disproportionate attention that compounds across layers~\citep{wu2025emergence}. While Mamba lacks attention, similar effects may emerge from recursive updates, enabling primacy-like behavior in deep models.

\subsection{Memory Biases in State Space Models}

State space models process sequences through recursive updates to a fixed-size hidden state, allowing them to operate in linear time. Unlike transformers, SSMs do not rely on token-to-token interactions, making them inherently biased toward recent inputs. This \textit{recency bias} has been formally analyzed in recent work on smoothing in state space models~\citep{wang2025understandingmitigatingbottlenecksstate,jelassi2024repeat}.

Surprisingly, \textit{primacy effects} have also been observed in SSMs. In particular, \citet{morita2025emergenceprimacyeffectstructured} shows that S4 models can exhibit strong primacy depending on how the discretization step \( \boldsymbol{\Delta}_t \) is initialized. In time-invariant SSMs such as S4, the values of \( \boldsymbol{\Delta}_t \) define the effective frequency spectrum the model can capture. This connection has been further explored in the context of autocorrelation-based perspectives, where SSMs are viewed as spectral filters~\citep{liu2024autocorrelation}.

A particularly illustrative case is the diagonal initialization used in S4D~\citep{gu2022parameterization}, where the continuous-time state matrix is initialized with complex numbers of the form $
\Lambda_n = -\frac{1}{2} + 2\pi i n,
$
for integer \( n \). When discretized, the eigenvalues become $
\lambda_n = \exp(\boldsymbol{\Delta}_t \Lambda_n),
$
which determines the poles of the system in the Z-domain. The magnitude of these poles governs the decay (i.e., memory retention), while their imaginary components define the resonant frequencies the model is sensitive to. Therefore, the choice of \( \boldsymbol{\Delta}_t \) directly modulates the system's frequency coverage and effective memory scale. In this view, short-term memory corresponds to responsiveness to high frequencies, while long-term memory reflects retention of low-frequency content.

\subsection{Time-Variant SSMs and Mamba}
 Unlike models like S4, where the parameters remain fixed over time, Mamba dynamically updates its state using input-dependent functions, including a learned discretization parameter \( \boldsymbol{\Delta}_t \). This breaks many of the assumptions used to analyze time-invariant models, making Mamba harder to interpret using standard signal-processing tools.

Although Mamba is time-variant, \( \boldsymbol{\Delta}_t \) plays a role similar to that in S4, acting as a learned forgetting gate. Smaller values of \( \boldsymbol{\Delta}_t \) correspond to longer memory and help identify components involved in long-term recall. Since \( \boldsymbol{\Delta}_t \) also appears in the input term, it controls how strongly new inputs are integrated. We further show that repeated or periodic inputs can bias \( \boldsymbol{\Delta}_t \), leading to faster forgetting. We explore this effect in Section~\ref{sec:semantic}.

%% file: sections/task.tex
\section{Model Architecture and Task}
\label{sec:task}

\subsection{Mamba Architecture}
Mamba is built on state space models (SSMs), which maintain structured hidden states \( \mathbf{h}_t \in \mathbb{R}^{d \times N} \) that evolve over time through recurrent updates. Each input dimension \( i \in \{1, \dots, d\} \) of the embedding vector \( \mathbf{x}_t \in \mathbb{R}^d \) is processed independently by a dedicated selective SSM block, enabling token-to-token interaction through the accumulation of information over time within the recurrent state dynamics. The state-update dynamics for each dimension are given by:

\begin{align}\nonumber
    \mathbf{h}_t^{(i)} &= \mathbf{A}_t^{(i)} \mathbf{h}_{t-1}^{(i)} + \mathbf{B}_t^{(i)} x_t^{(i)}, \\
    y_t^{(i)} &= \mathbf{C}_t \mathbf{h}_t^{(i)} + D^{(i)} x_t^{(i)},
    \label{eq:ssm_dynamics}
\end{align}
where:
\begin{itemize}
    \item \( \mathbf{h}_t^{(i)} \in \mathbb{R}^{N} \) is the hidden state for dimesion \( i \),
    \item \( \mathbf{A}_t^{(i)} \in \mathbb{R}^{N \times N} \) is the recurrence matrix or the forget gate (diagonal in Mamba; possibly full or low-rank in models such as S4\footnote{In S4D \cite{gu2021efficiently}, \( \mathbf{A} \) is diagonal with entries initialized in the complex plane; \( \mathbf{B} \) and \( \mathbf{C} \) are also complex-valued, and are discretized via the bilinear transform. These variants differ in expressivity, stability, and frequency selectivity.}),
    \item \( \mathbf{B}_t^{(i)} \in \mathbb{R}^{N} \), \( \mathbf{C}_t \in \mathbb{R}^{N} \) are the input and output projection vectors (gates), respectively. \( D^{(i)} \in \mathbb{R} \) is a residual weight,
    \item \( x_t^{(i)} \in \mathbb{R} \) is the \( i \)-th dimension of the input at time \( t \).
\end{itemize}

\vspace{0.3em}
These models can be interpreted as implicit convolutional filters applied in the frequency domain, where long-range dependencies are managed through kernel design rather than explicit attention. To enhance adaptability, Mamba introduces a learned per-token discretization parameter \( \boldsymbol{\Delta}_t \), which dynamically modulates the recurrence behavior at each time step. This allows the model to adjust the rate at which information is integrated or forgotten, depending on the input structure—a property central to our later analysis of memory effects. We conduct our experiments primarily on \emph{Falcon Mamba 7B}~\citep{zuo2024falconmambacompetitiveattentionfree}, and validate our findings on \emph{Mamba 1.4B}~\citep{gu2023mamba} to ensure consistency across model variants.


\subsection{Structured Recall Task}

To investigate how Mamba allocates memory across time, we design a structured recall task inspired by serial position experiments in cognitive psychology. Each input sequence consists of $L$ subject–relation–object (\texttt{s–r–o}) triples, followed by a query targeting the object at a specific position in the sequence. The input fed to the model is structured as follows:

\begin{tcolorbox}[colback=gray!5!white, colframe=black!70, title=Input format]
\textbf{Context}: \texttt{s\textsubscript{1} r\textsubscript{1} o\textsubscript{1}}. 
\texttt{s\textsubscript{2} r\textsubscript{2} o\textsubscript{2}}. $\cdots$ 
\texttt{s\textsubscript{L} r\textsubscript{L} o\textsubscript{L}}.

\textbf{Query}: \texttt{s\textsubscript{k} r\textsubscript{k}}
\end{tcolorbox}
The model is expected to generate the correct \texttt{object\textsubscript{k}} corresponding to the $k$-th triplet in the original sequence. To ensure consistent tokenization, we construct the dataset using subject, relation, and object terms that are each represented by a single unique token in the model's vocabulary. Additionally, we ensure that all subjects and objects within a sequence are distinct and do not repeat. For each target position $k$, we generate 50 unique input sequences and report the average accuracy, allowing a detailed analysis of the recall performance of the model in different sequence positions. The details of the dataset construction can be seen in the Appendix \ref{sec:Dataset Contstruction}.

To better understand the conditions under which memory biases emerge, we vary two key factors:

\begin{enumerate}
    \item \textbf{Sequence Depth ($L$):} We test sequences of varying lengths ($\{8, 16, 32, 64, 128\}$) to examine how recall performance changes with context size and whether memory degrades with longer input.
    
    \item \textbf{Relation Type:} We compare two variants: (i) \textbf{Repeated Relation}, where all relation tokens are identical (e.g., \texttt{Mike} \texttt{likes} \texttt{Sara.} \texttt{Olivier} \texttt{likes} \texttt{Alice.} \texttt{...} \texttt{. Mike} \texttt{likes}), introducing structural repetition in the input; and (ii) \textbf{Random Relation}, where relations are sampled uniquely from a fixed vocabulary, introducing input diversity(e.g \texttt{Mike} \texttt{loves} \texttt{Sara.} \texttt{Olivier} \texttt{hates} \texttt{Alice.} \texttt{...} \texttt{. Mike} \texttt{loves}).
\end{enumerate}

\begin{figure}[t]
  \centering
  \includegraphics[width=\columnwidth]{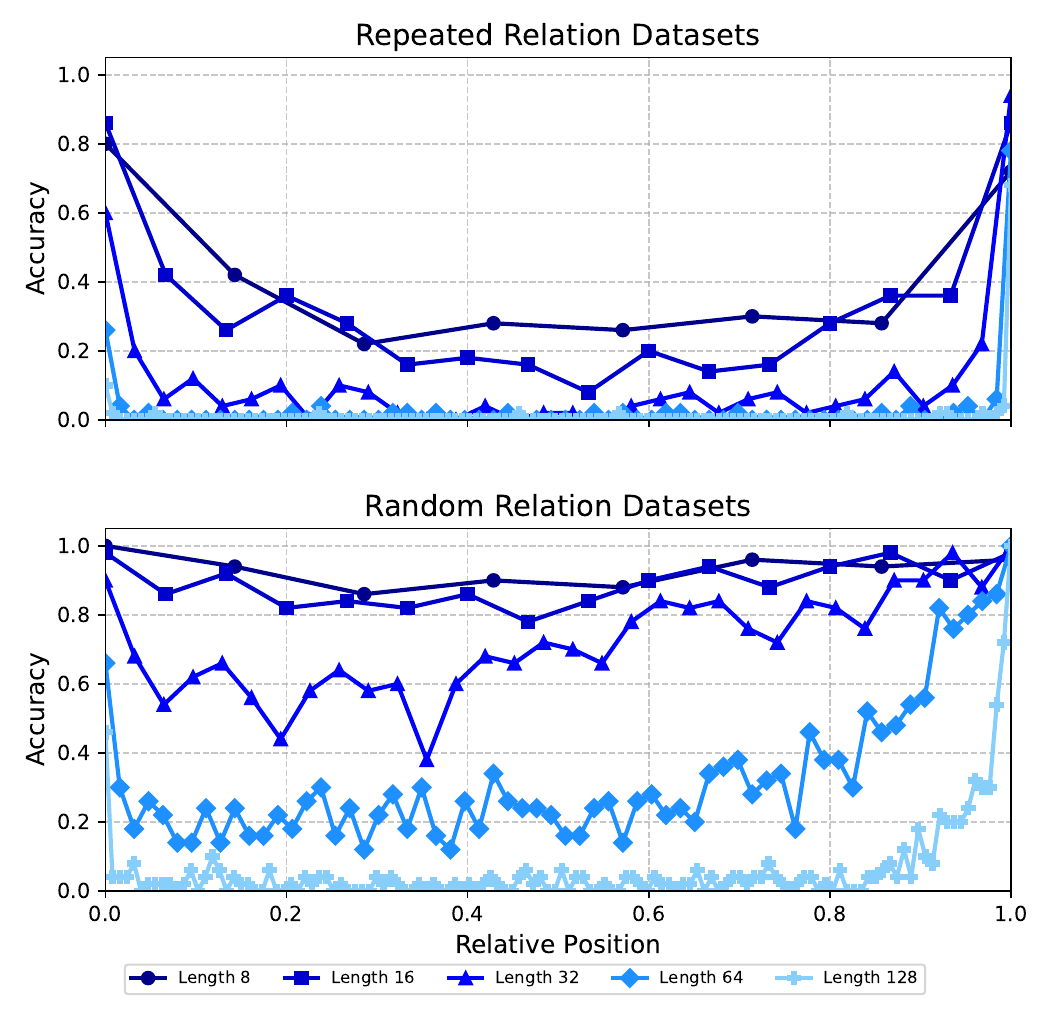}
  \caption{Recall accuracy of Falcon Mamba 7B as a function of position $k$ in the input sequence. Accuracy peaks at the beginning and end, forming a U-shaped curve characteristic of primacy and recency effects.}
  \label{fig:U-shape}
\end{figure}

%% file: sections/results.tex
\section{Experimental Results and Findings}
\label{sec:results}
\subsection{The Emergence of Primacy and Recency Effects}

Across variations in both sequence depth and relation type, we consistently observe a U-shaped recall accuracy curve: the model achieves higher accuracy at early and late positions, with a notable dip in the middle. This pattern mirrors the classic primacy and recency effects observed in human memory, as shown in Figure~\ref{fig:U-shape}. Interestingly, the shape of this curve depends not only on position but also on input semantics. For example, the random relation setting shows better performance in the middle positions compared to the repeated one. In contrast, the repeated relation condition exhibits a more pronounced U-shape, with a sharper drop in accuracy at central positions. This suggests that structural regularity in the input, which is introduced by repeating the same relation token, may lead the model to overweight positional edges, reinforcing primacy and recency effects. 
\begin{figure*}[t]
  \centering
  \begin{subfigure}[t]{0.48\textwidth}
    \centering
    \includegraphics[width=\linewidth]{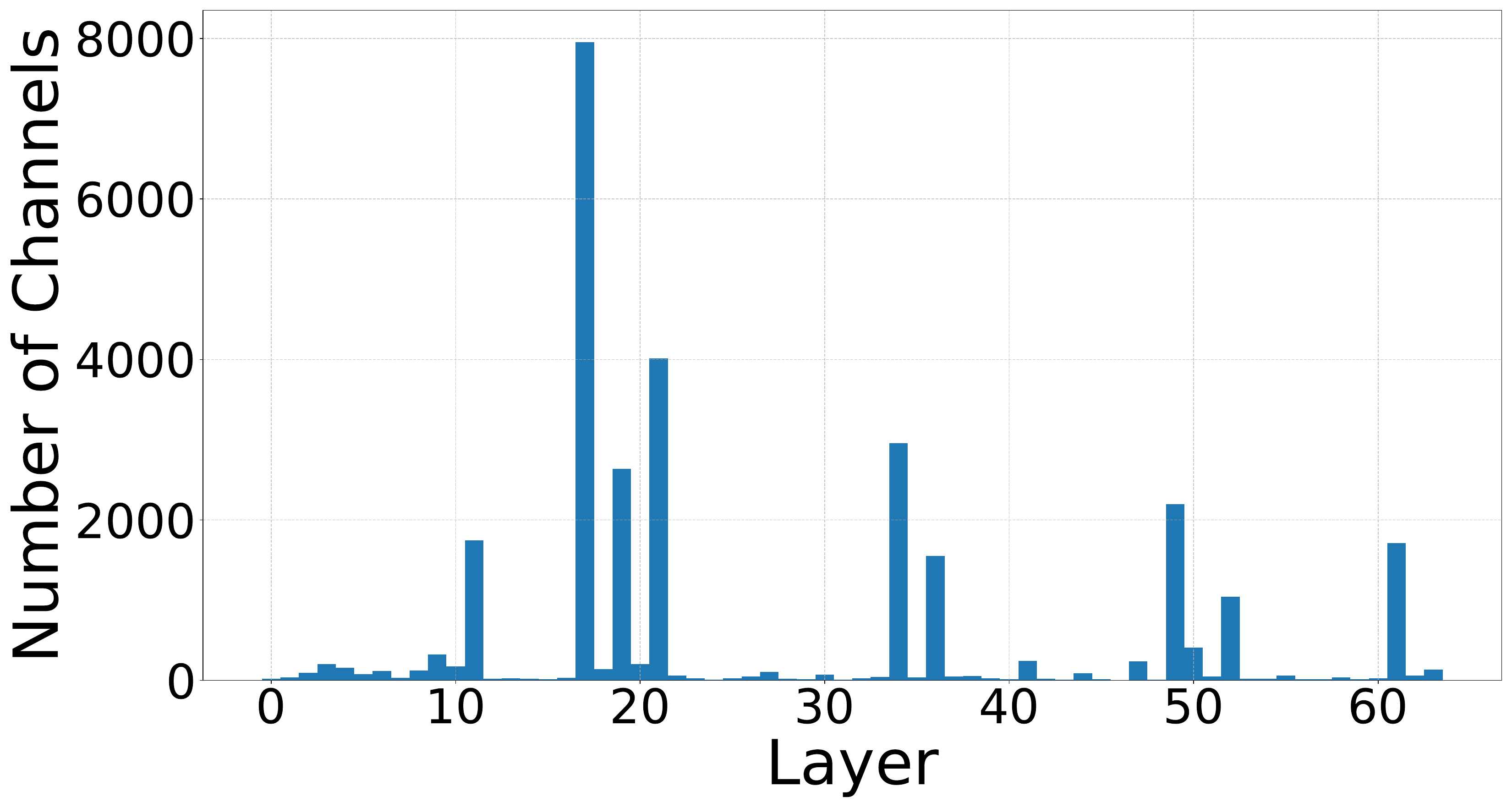}
    \caption{}
    \label{fig:LTM}
  \end{subfigure}
  \hfill
  \begin{subfigure}[t]{0.48\textwidth}
    \centering
    \includegraphics[width=\linewidth]{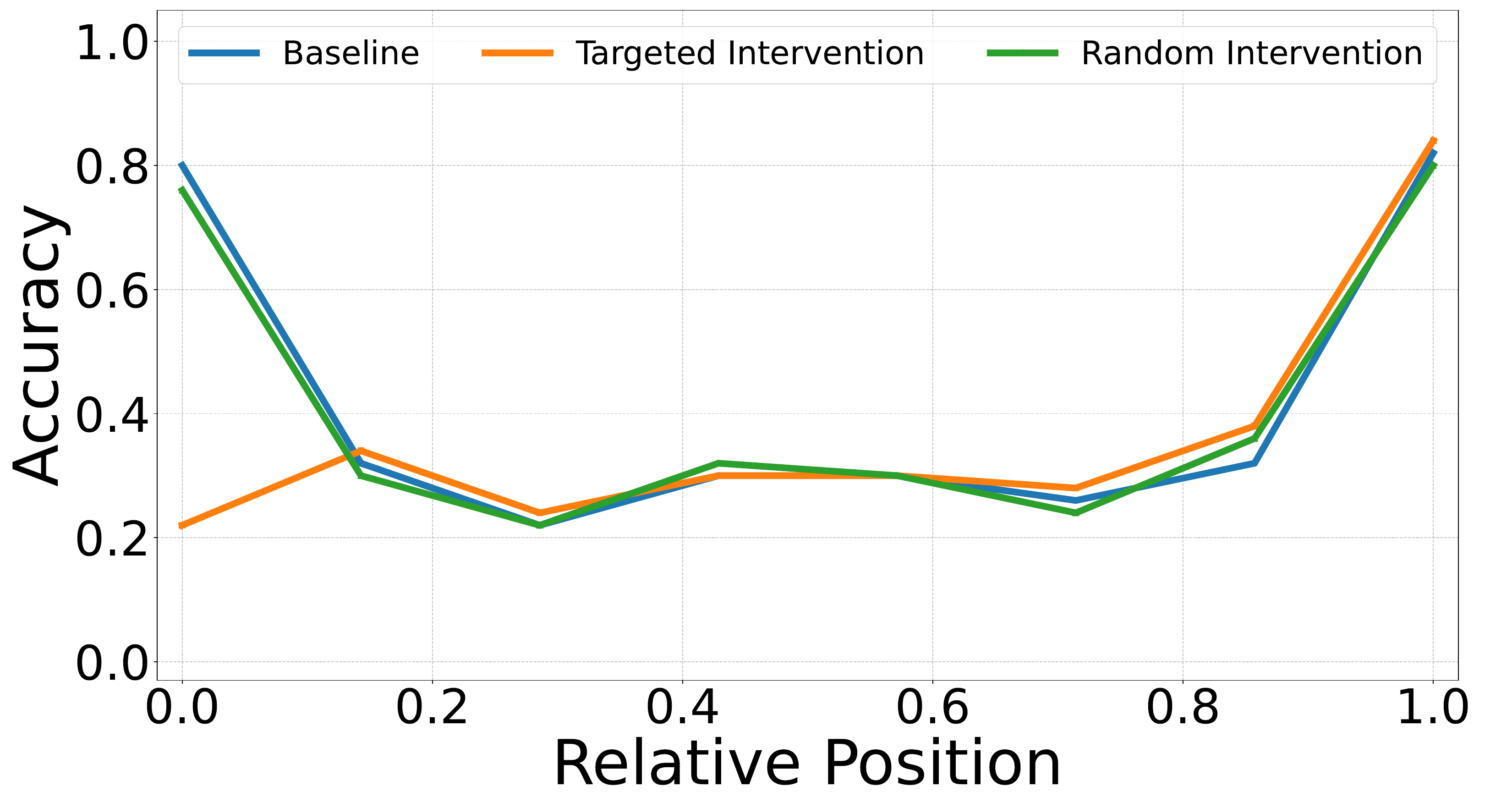}
     \caption{}
    \label{fig:intervention}
  \end{subfigure}
  \caption{Layer-wise organization and functional role of long-term memory channels in Falcon Mamba 7B. \textbf{(a)} Distribution of long-term memory channels ($P^{(i)} > 0.7$ with $\tau = 0.7$) across Falcon Mamba 7B layers.  \textbf{(b)} First-position recall drops sharply when intervening on the state matrix of identified long-memory channels, confirming their role in retaining early input information.}

  \label{fig:ltm+intervention}
\end{figure*}
These observations motivate a deeper investigation into the underlying mechanisms driving these effects and raise three fundamental questions that we explore in the following subsections:

\begin{itemize}
    \item \textbf{Why Primacy?} Does early input persist due to long-term information retention in a specialized long memory? Where such memory is located?
    \item \textbf{Why Recency?} Is the preference for recent inputs a consequence of architecture bias or memory decay? 
    \item \textbf{Why Semantic Dependence?} How does input regularity (e.g., repeated relations triplets) interact with input- and forget-gates, to shape recall behavior?
\end{itemize}

\subsection{Why Primacy? The Existence of "Long-Term Memory"}
\label{sec:primacy}
Inspired by the multi-store memory model in cognitive psychology~\cite{glanzer1966two}, we hypothesize that certain channels in Mamba's selective SSM blocks serve as long-term memory units. These channels are capable of maintaining influence from early inputs over long contexts.

Focusing on the recurrence from Equation~\ref{eq:ssm_dynamics}, and assuming standard initialization \( (\mathbf{h}_0^{(i)} = 0 \)), the unrolled state dynamics yield:

\begin{equation}
    \mathbf{h}_t^{(i)} = \sum_{j=1}^{t} \left( \prod_{k=j+1}^{t} \mathbf{A}_k^{(i)} \right) \mathbf{B}_j^{(i)} x_j^{(i)}.
\end{equation}



We propose a quantitative test for long-term memory retention based on the magnitude of early input contribution. For each channel \( i \), we compute the memory coefficient:

\begin{equation}
    \mathcal{M}^{(i)} := diag(\prod_{t=2}^{T} \mathbf{A}_t^{(i)}) \in [0,1]^{N} 
\end{equation}
where \( T \) is the final timestep in the context (prior to the query token). We define the long-term memory probability of the channel $i$ given threshold $\tau\in[0,1]$ as:

\begin{equation}
P^{(i)}(\tau) := \frac{\left| \forall j \in [N], \ \mathcal{M}^{(i)}_j \geq \tau \right|}{N},
\end{equation}
where $[N]:=\{1,2,\dots,N\}$. 



\begin{tcolorbox}[colback=gray!5!white, colframe=black!70, title=Definition: Long-Term Memory]
We say that channel \( i \) exhibits \emph{long-term memory with probability \( p\in [0,1] \)} if:

\[
P^{(i)}(\tau) > p,
\]

\end{tcolorbox}
The complete identification pipeline is summarized in Algorithm~\ref{alg:contribution-filter} in the Appendix.

Using the procedure described above with threshold $\tau=0.7$ and probability cut-off $p=0.7$, we identify channels exhibiting long-term memory behavior. As shown in Figure~\ref{fig:LTM}, these channels are not uniformly distributed but instead concentrate on specific layers of the Mamba architecture. In particular, Layer 17 stands out with a disproportionately high number of long-memory channels, suggesting a specialized role in preserving early input information.

\paragraph{Causal intervention:}
To investigate whether identified channels actively contribute to long-term memory, we performed a controlled intervention based on forget gates $\mathbf{A}_t$, which regulate memory in Mamba. Specifically, we ablate the gates corresponding to the channels with high long-term memory scores $P^{(i)}(0.7)>0.7$ as identified earlier, by targeting the three layers with the highest density of long-memory channels (see next paragraph for detailed ablations). This intervention zeroes out the recurrence matrices $\mathbf{A}_t$ at the timestep of the first triplet, effectively blocking early information from entering these long-memory pathways.

To validate the specificity of this effect, we compare it to a baseline where an equal number of randomly selected channels are ablated. As shown in Figure~\ref{fig:intervention}, only the targeted intervention leads to a significant drop in recall accuracy for the first position, confirming the functional importance of these channels. In contrast, random ablation leaves performance largely intact. These results demonstrate that long-term memory is not uniformly spread but is localized in structurally specialized dimensions.

\begin{figure}[h]
  \centering
  \includegraphics[width=\columnwidth]{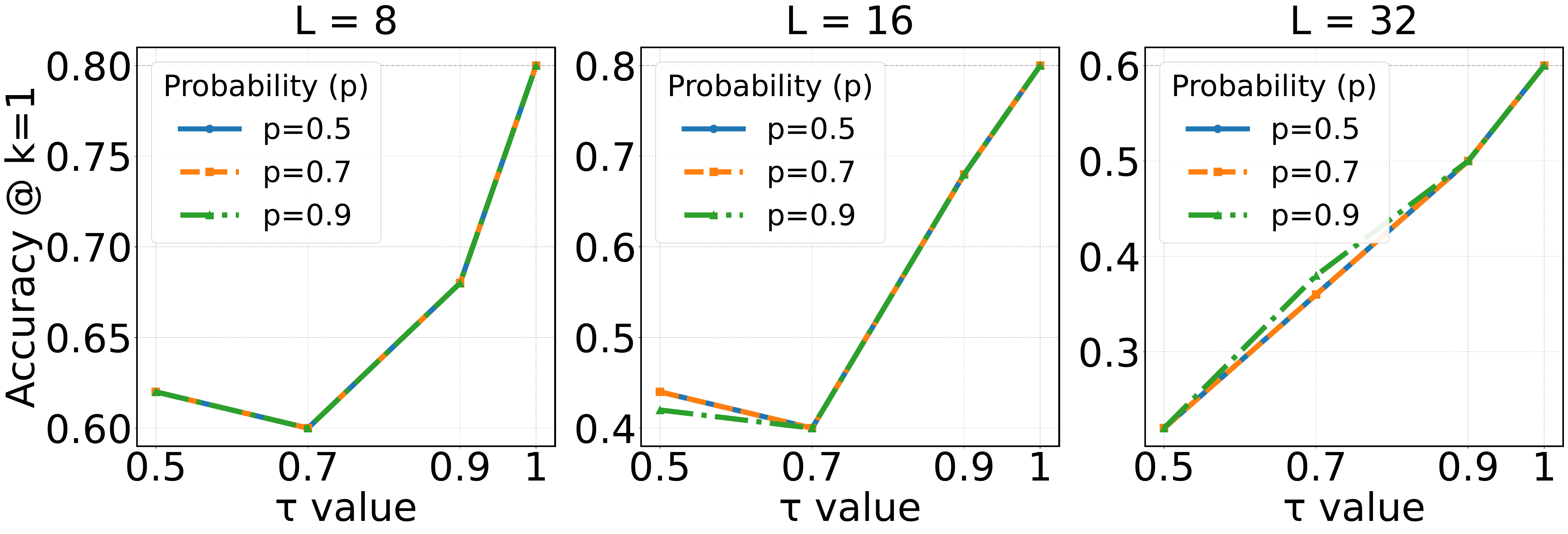}
  \caption{Impact of \(\tau\) and $p$ on first-position recall under long-term memory interventions. High-precision selection (e.g., $p=0.9$,\(\tau=0.7\)) reliably degrades early recall, validating our identification method. The effect intensifies with longer sequences, while lower \(\tau\) values recruit more channels, reflecting increased memory demands.}
  
  \label{fig:comparison}
\end{figure}

\paragraph{The choice of \(p\) and \(\tau\):}
To evaluate the sensitivity of our long-term memory identification method, we conducted an ablation study varying the two key parameters: the threshold \( \tau \), which controls the strictness of what constitutes significant memory retention, and \( p \), the proportion of states exceeding this threshold. Notably, when \( \tau = 1 \), no states qualify. We experimented with combinations of \( p \in \{0.5, 0.7, 0.9\} \) and \( \tau \in \{0.5, 0.7, 0.9\} \), and report the impact of interventions targeting the top-1 layer identified under each criterion (Figure~\ref{fig:comparison}).

The results show a clear trend: high-precision selection of long-term memory channels (e.g., \( p = 0.9 \)) with a moderately strict threshold (\( \tau = 0.7 \)) is already sufficient to yield a significant drop in recall accuracy at the first position, demonstrating the effectiveness of our identification procedure. Additionally, as shown in the rightmost Figure~\ref{fig:comparison}, the degradation in recall becomes more pronounced as the input sequence length increases, indicating that longer contexts rely more heavily on the identified long-term channels. Finally, we observe that \( \tau \) decreases with sequence length, suggesting that a larger number of channels become involved in preserving long-range information under relaxed selection criteria.


\paragraph{Initialization of state:}  
We investigated an alternative scheme for initializing the recurrent state, moving away from the default setting \( \mathbf{h}_0^{(i)} = 0 \). In this setup, we initialized \( \mathbf{h}_0^{(i)} \) from a uniform distribution \( \mathcal{U}(0, 1) \), one layer at a time. This change introduces non-zero pre-activation signals, which may influence the temporal dynamics of the model by modifying how early inputs are processed and retained.

\begin{figure}[t]
  \centering
  \includegraphics[width=\columnwidth]{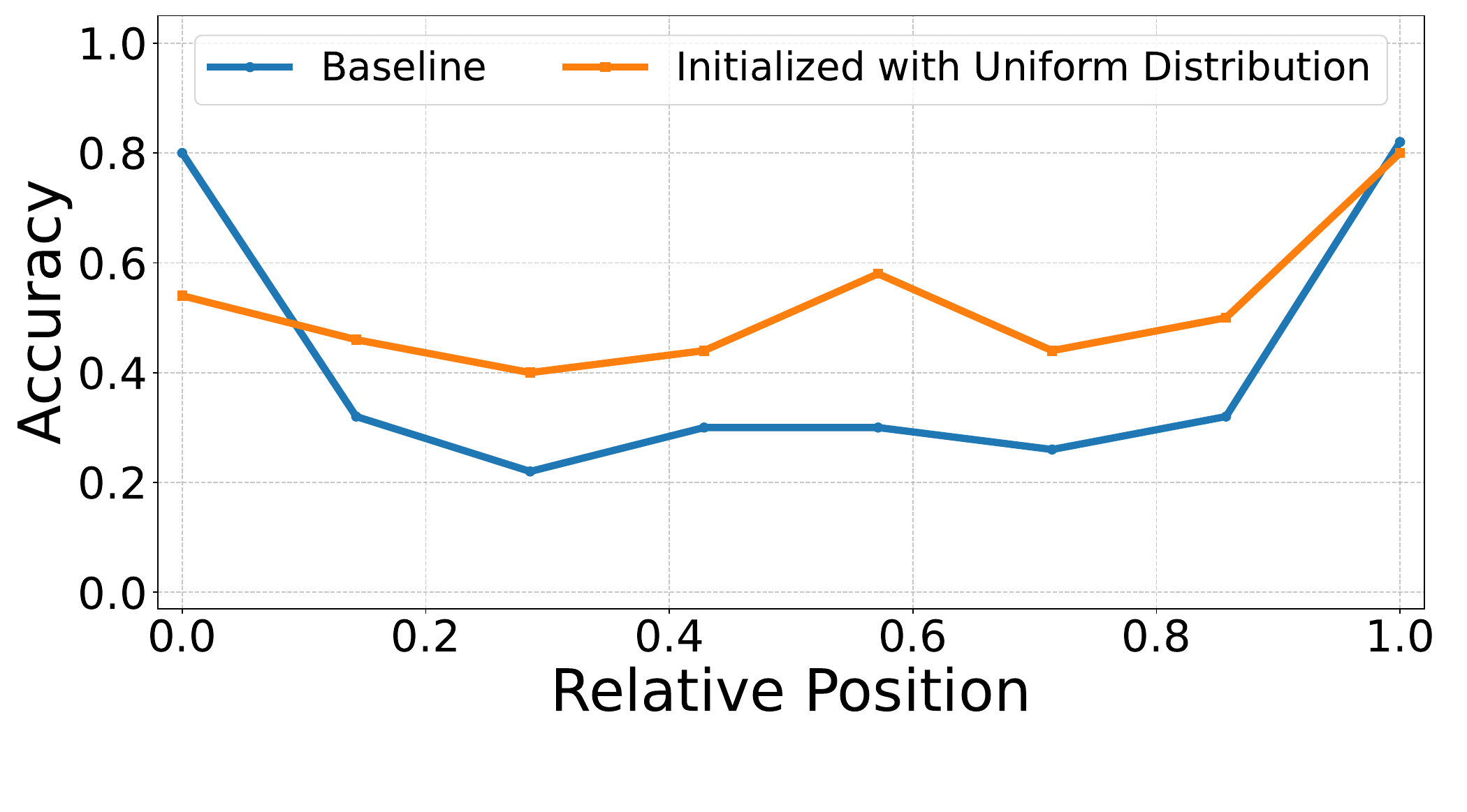}
  \caption{Effect of initializing the recurrent state at Layer 31 using uniform values on repeated relation}
  \label{fig:hiddenInit}
\end{figure}

Figure~\ref{fig:hiddenInit} illustrates the effect of modifying Layer 31's initialization on recall accuracy across different relative positions in the repeated relation setting. Recall accuracy primarily at middle positions gets increased, mitigating the “lost in the middle” effect and flattening the U-shaped curve. This suggests that the zero-initialized state may disproportionately favor early positions.

\subsection{Why Recency? Dynamics and Short-Term Memory}
\label{sec:recency}
Recency is the most consistent effect observed in our structured recall task. Performance near the end of the sequence is systematically higher than the middle, even under conditions of input randomness (e.g., random relation setting).

We interpret this through the lens of Mamba’s update rule: newer inputs have had fewer decay steps applied, yielding a kind of implicit recency gating. This aligns with the classic \emph{two-store model} of human memory \cite{glanzer1966two}.

\paragraph{Exponential decay intuition.}
Following the S6 analysis in \citet{gu2023mamba}, let the output at time $N$ be:

\begin{equation}\label{eqn:contribution}
    {y}_t^{(i)} = \mathbf{C}_t\sum_{j=1}^{t} \underbrace{\left( \prod_{k=j+1}^{t} \mathbf{A}_k^{(i)} \right) \mathbf{B}_j^{(i)}}_{j^{th}-contribution} x_j^{(i)}.
\end{equation}
Clearly, the contribution of the $j^{th}$ inputs is proportional to $ A^{t-j}
$. In other words, inputs dominate exponentially with time. This is investigated as well in earlier studies~\cite{wang2025understandingmitigatingbottlenecksstate}.

\paragraph{Ablation and distraction experiment.}
We empirically validate this dynamic by inserting random distractor tokens of varying lengths between the main sequence and the query prompt, as follow. 

\begin{figure}[t]
  \includegraphics[width=\linewidth]{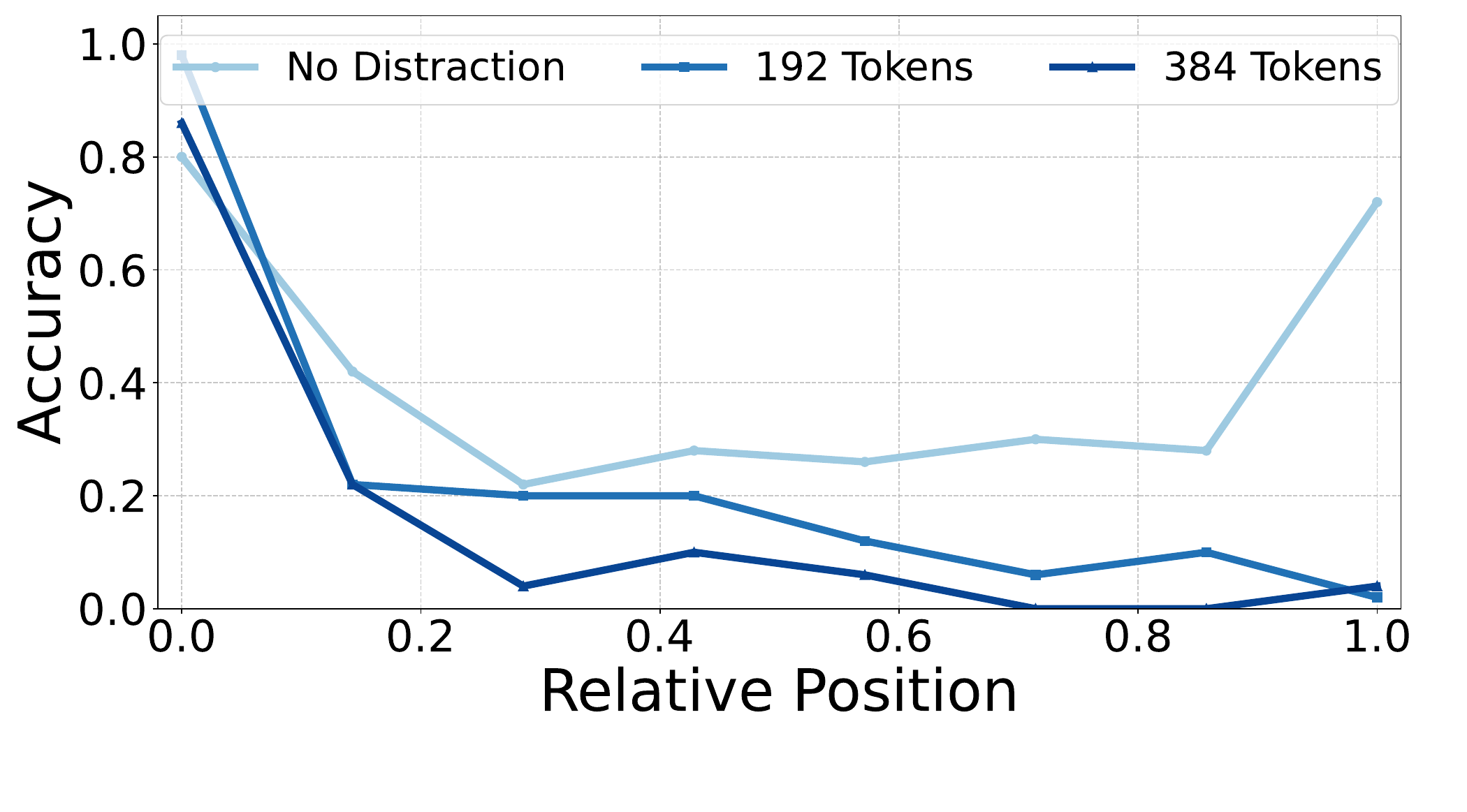}
  \caption{Adding distraction tokens disrupts the accuracy of the recall, especially in the recent positions}
  \label{fig:recency}
\end{figure}

\begin{tcolorbox}[colback=gray!5!white, colframe=black!70, title=Input format]
\textbf{Context}: \texttt{s\textsubscript{1} r\textsubscript{1} o\textsubscript{1}}. 
\texttt{s\textsubscript{2} r\textsubscript{2} o\textsubscript{2}}. $\cdots$ 
\texttt{s\textsubscript{L} r\textsubscript{L} o\textsubscript{L}}.

\textbf{Query}: \texttt{$n$-random tokens} \texttt{s\textsubscript{k} r\textsubscript{k}}
\end{tcolorbox}
As shown in Figure~\ref{fig:recency}, recall accuracy degrades consistently across all positions as the number of random tokens \(n\) increases. This suggests that recency is not an absolute feature of the architecture, but is dynamically sustained by state transitions that saturate over time, a behavior characteristic of short-term memory.

\subsection{Why Semantic Dependence? Input Periodicity and Delta Dynamics}
\label{sec:semantic}

In addition to positional effects, our structured recall task reveals a third axis of variation: semantic regularity. Sequences with repetitive structure (e.g. repeated relations) show stronger U-shaped recall compared to random sequences. We hypothesize that this is due to Mamba’s adaptive memory mechanism, specifically the behavior of its learned discretization factor \( \boldsymbol{\Delta}_t \).



In Mamba, the per-token gate \( \boldsymbol{\Delta}_t \) governs memory dynamics by controlling state updates. Computed via a low-rank projection and \texttt{softplus} activation on \( \mathbf{x}_t \), it determines how quickly past information decays, where larger values induce fast forgetting, while smaller values preserve memory.

Importantly, \( \boldsymbol{\Delta}_t \) modulates both the internal state transition matrix \( \mathbf{A}_t \) and the input scaling matrix \( \mathbf{B}_t \), thereby influencing how past information is forgotten and how current input is integrated. Channels with smaller \( \boldsymbol{\Delta}_t \) values result in slower decay within \( \mathbf{A}_t \), supporting long-term memory retention. In contrast, larger \( \boldsymbol{\Delta}_t \) values correspond to faster forgetting and stronger immediate input influence through \( \mathbf{B}_t \). 



To test how delta responds to input regularity, we construct synthetic sequences of length 64 with controlled repetition frequencies. From a small vocabulary, we sample a token and repeat it every \( k \in \{2, 4, 8, 16, 32, 64\} \) positions, where \( k=64 \) corresponds to a fully random (non-repetitive) sequence. These inputs are passed to Mamba, and we record the delta values across time and layers.

\begin{figure}[h]
    \centering
    \includegraphics[width=0.84\linewidth]{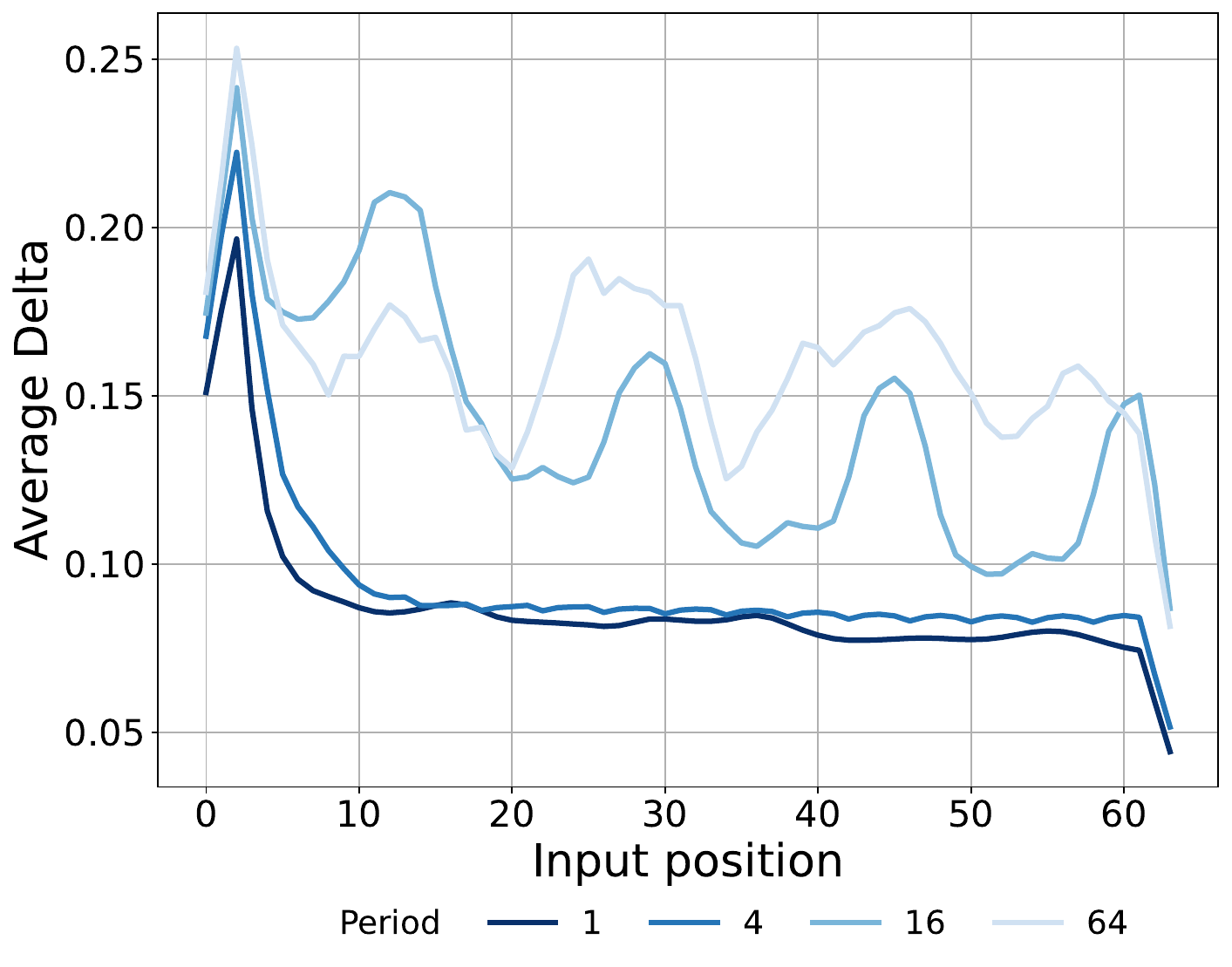}
    \caption{Average $\Delta$ across layers and channels for inputs with varying periodicities. Lower-frequency (longer-period) inputs induce larger $\Delta$ values, reflecting stronger integration of inputs.}
    \label{fig:delta-lines}
\end{figure}

\paragraph{Global Trends: Average Delta across Layers.}
Figure~\ref{fig:delta-lines} shows the delta values averaged across all layers and channels. We observe a clear monotonic relationship: sequences with longer repetition periods (i.e., lower frequency) result in higher average delta values. This suggests that Mamba allocates more input responsiveness when the input varies slowly, while rapidly changing inputs (high-frequency repetition, e.g., \( k=2 \)) suppress delta, leading to small memory decay.

\begin{figure}[ht]
    \centering

    \begin{subfigure}[t]{0.45\linewidth}
        \includegraphics[width=\linewidth]{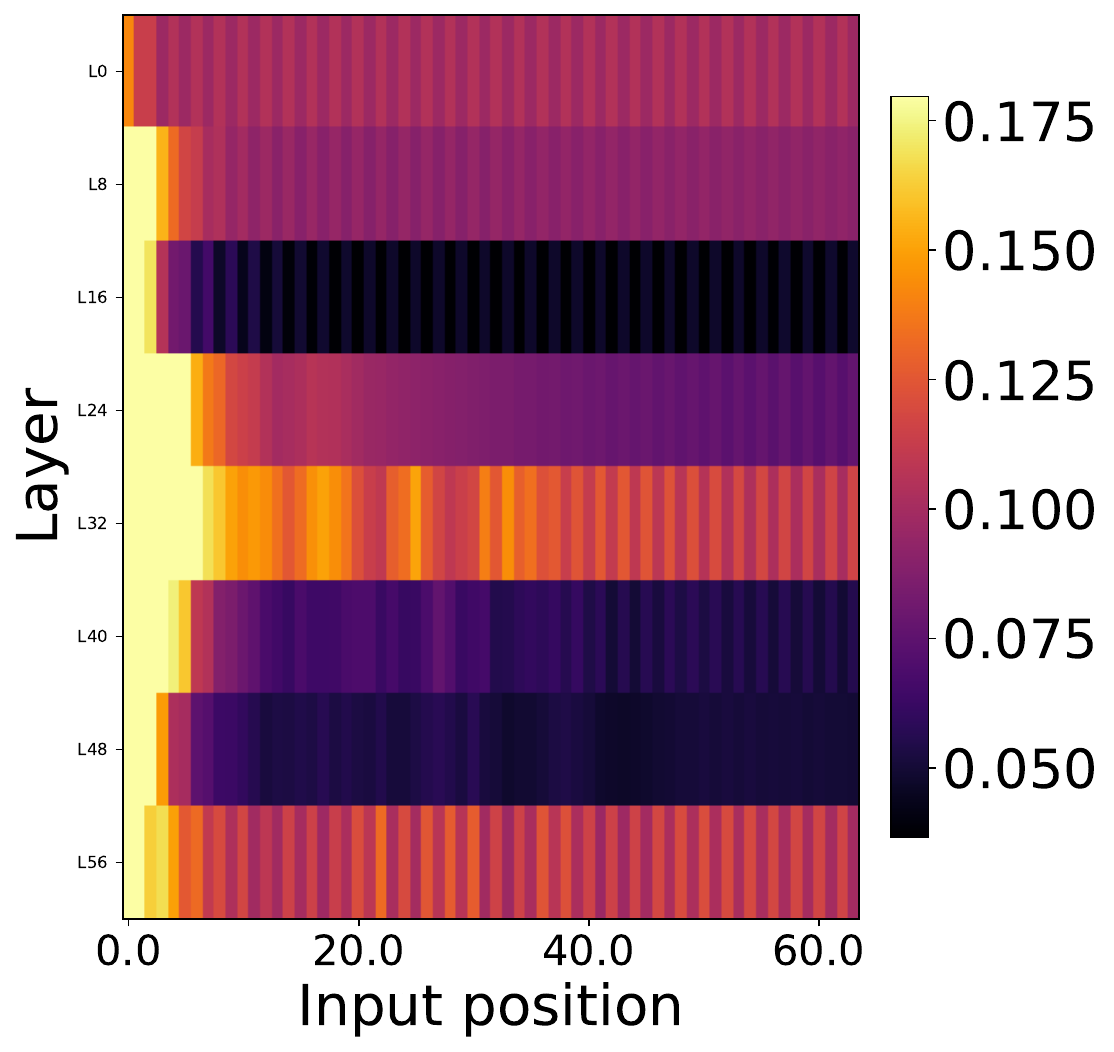}
        \caption{Period 2}
    \end{subfigure}
    \hfill
    \begin{subfigure}[t]{0.45\linewidth}
        \includegraphics[width=\linewidth]{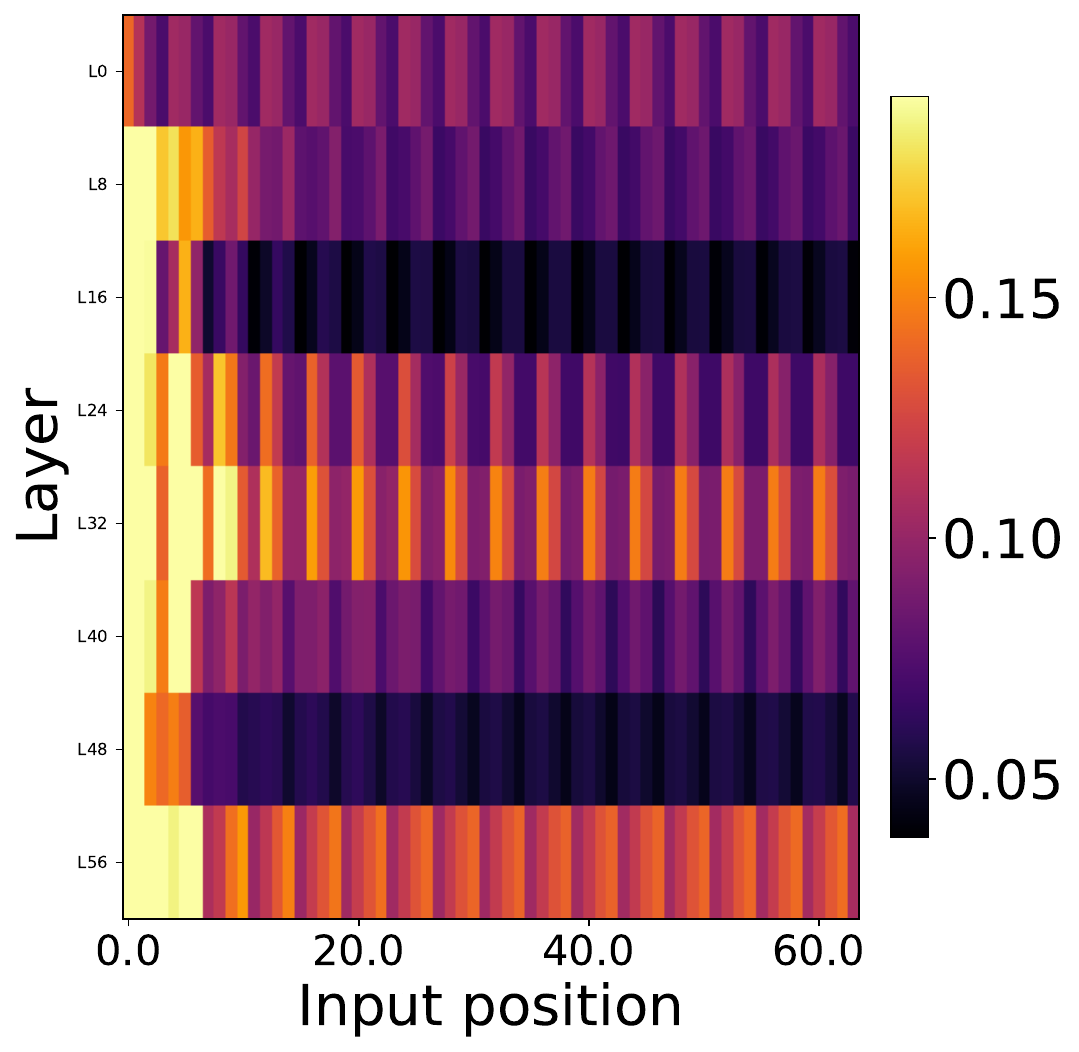}
        \caption{Period 4}
    \end{subfigure}

    \vspace{1em}

    \begin{subfigure}[t]{0.45\linewidth}
        \includegraphics[width=\linewidth]{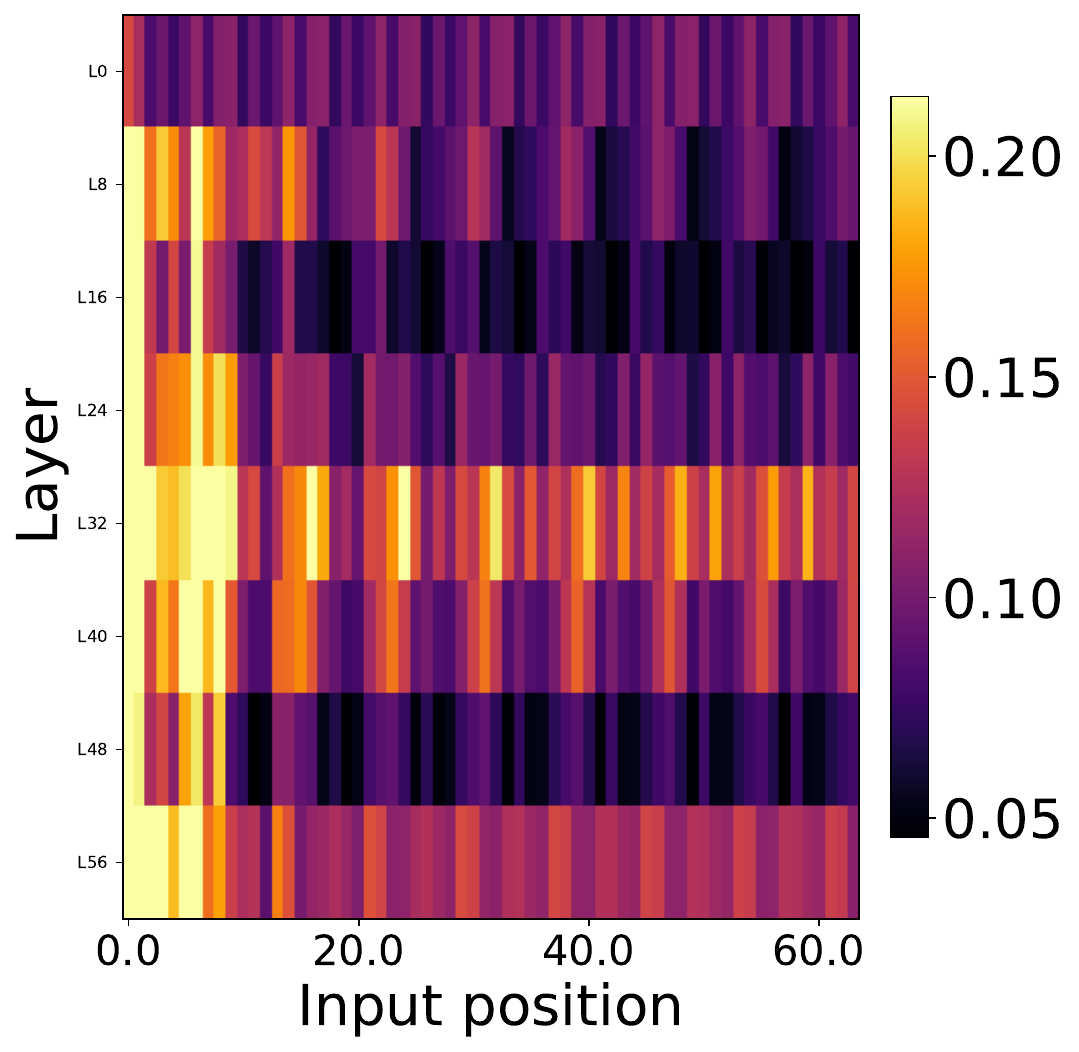}
        \caption{Period 8}
    \end{subfigure}
    \hfill
    \begin{subfigure}[t]{0.45\linewidth}
        \includegraphics[width=\linewidth]{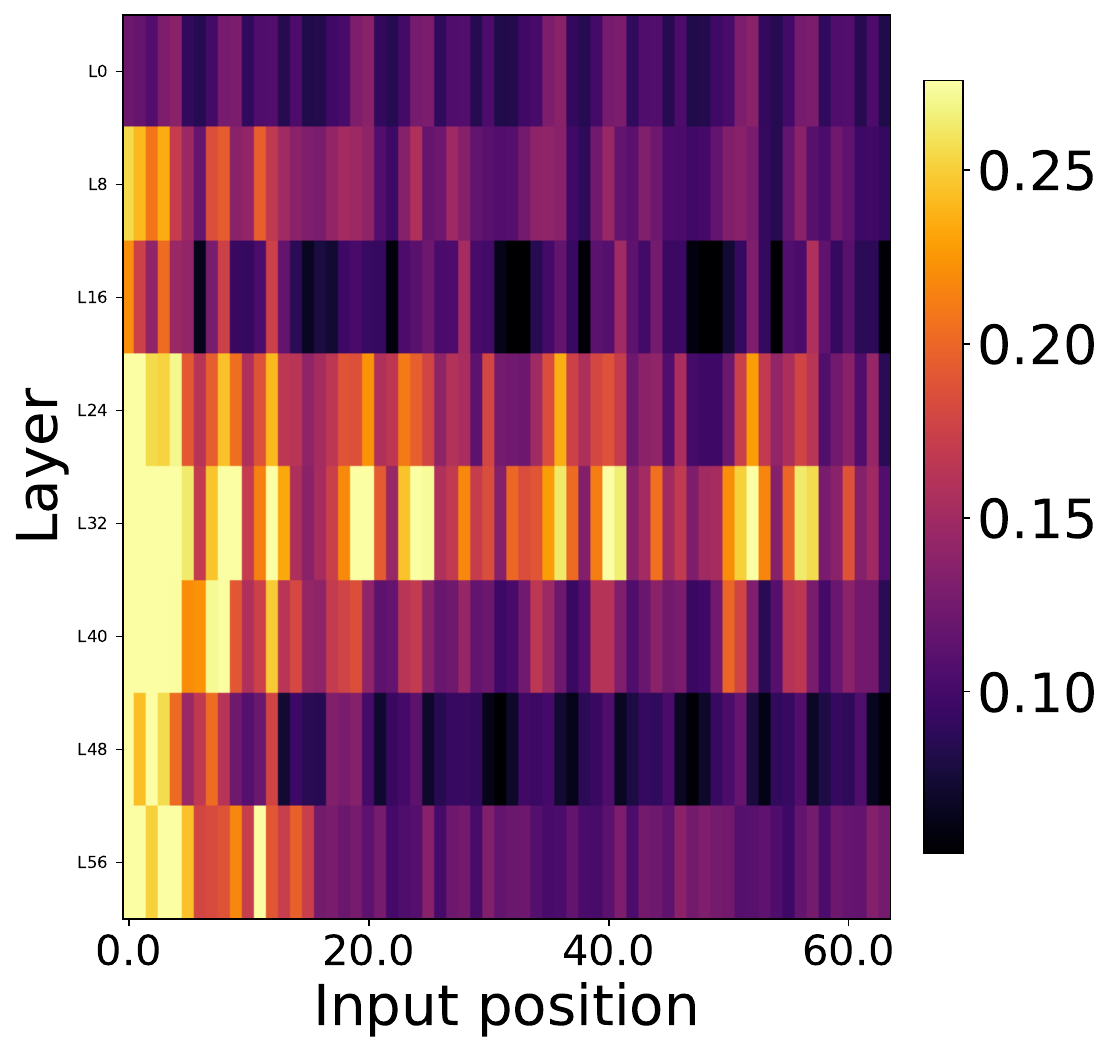}
        \caption{Period 16}
    \end{subfigure}

    \vspace{1em}

    \begin{subfigure}[t]{0.45\linewidth}
        \includegraphics[width=\linewidth]{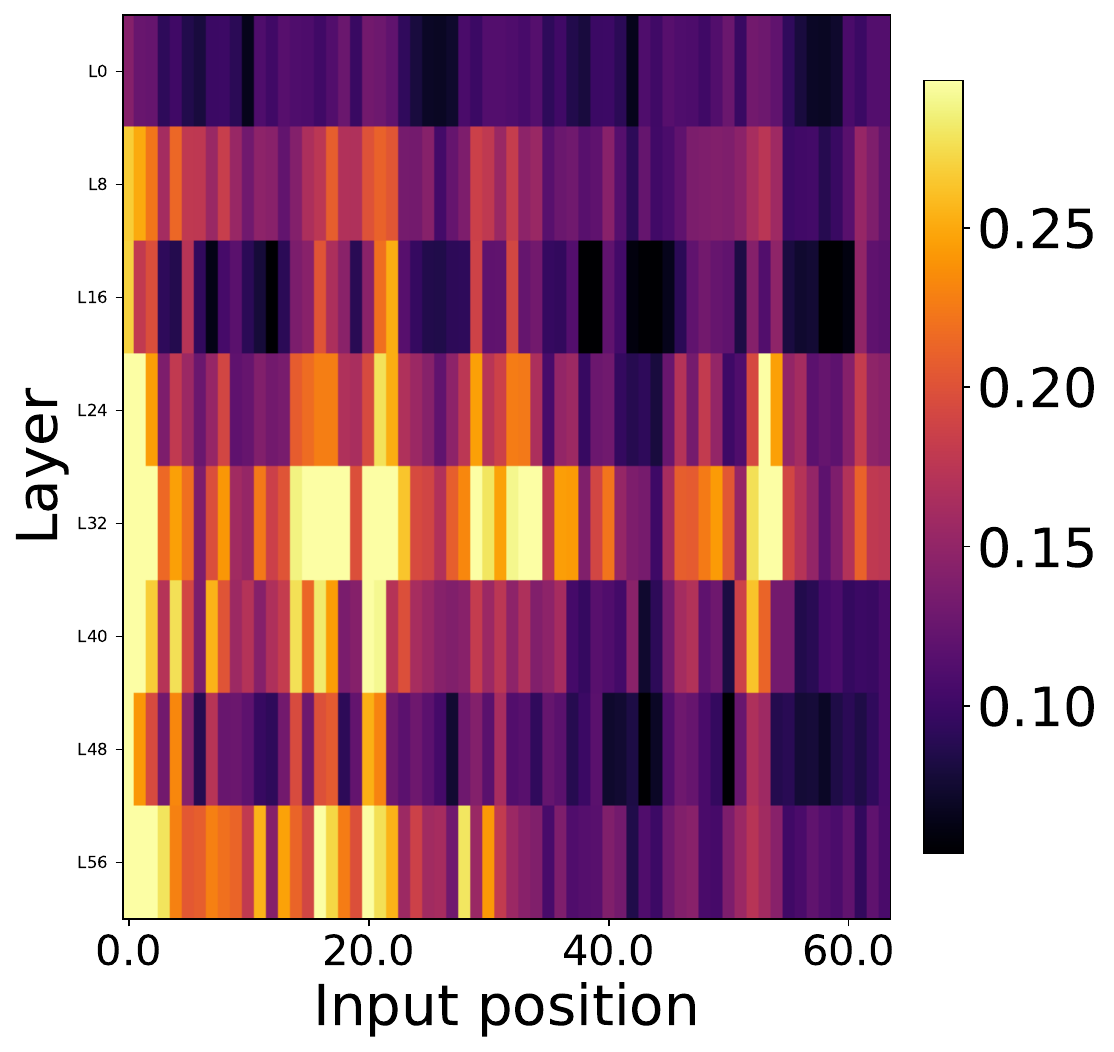}
        \caption{Period 32}
    \end{subfigure}
    \hfill
    \begin{subfigure}[t]{0.45\linewidth}
        \includegraphics[width=\linewidth]{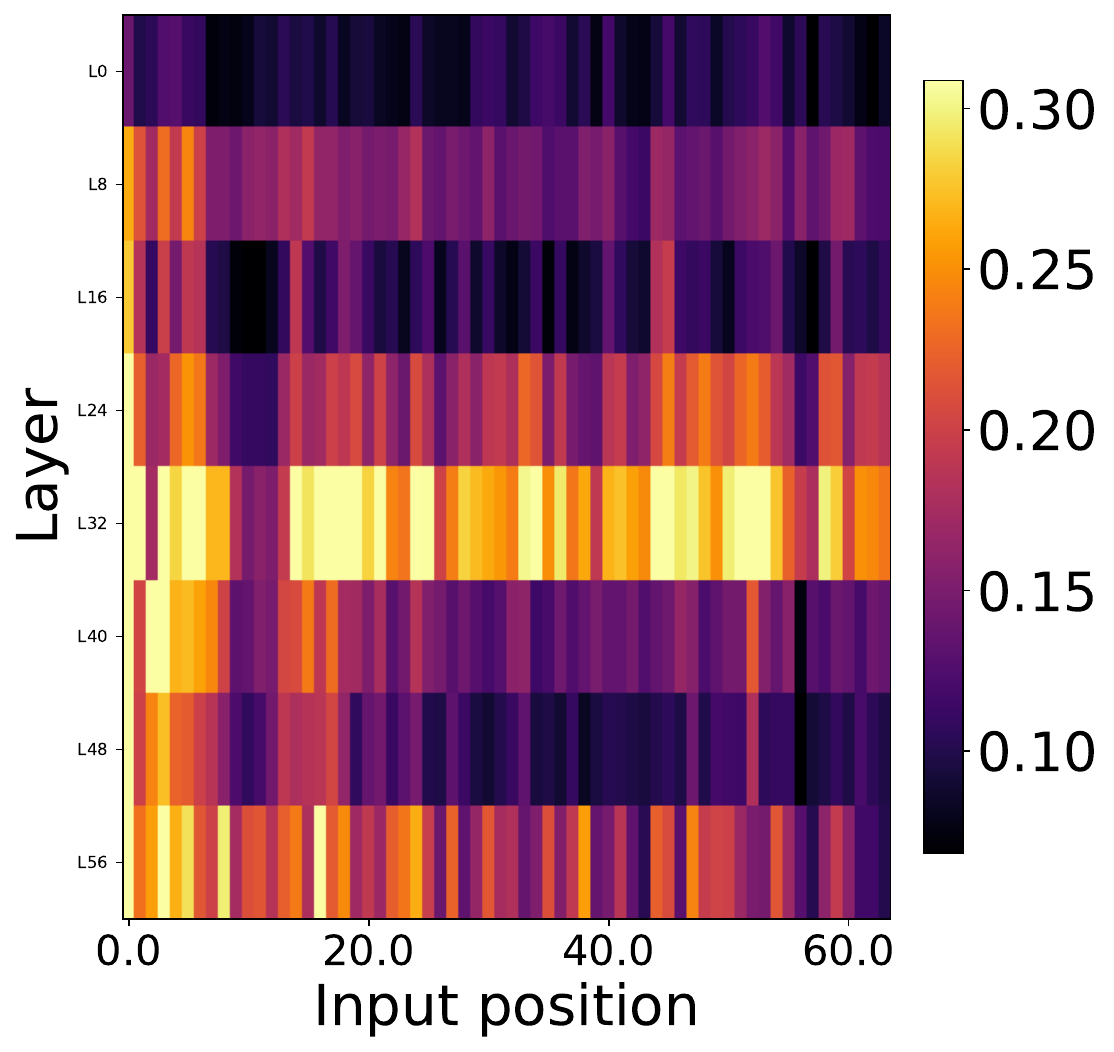}
        \caption{Period 64}
    \end{subfigure}

    \caption{Average $\Delta$ per layer across input positions for periodic inputs. Early layers show strong correlation with input frequency: lower-frequency (longer-period) patterns yield smaller $\Delta$, indicating slower forgetting.}
    \label{fig:avg-delta-periods}
\end{figure}

\paragraph{Layer-wise Delta Heatmaps.}
To understand how this trend evolves across depth, we plot delta values for each layer averaged over channels (Figure~\ref{fig:avg-delta-periods}). In early layers, delta shows temporal repetition aligned with the input frequency, but without strong modulation in magnitude. As we go deeper, two patterns emerge:
\begin{itemize}
    \item Repetition in delta values persists across time, reflecting sensitivity to periodic structure.
    \item The scale of delta values increases with layer depth, particularly for longer-period sequences, indicating that deeper layers progressively reactivate input injection (via higher delta) when input frequency is low.
\end{itemize}

\begin{figure}[t]
  \includegraphics[width=\linewidth]{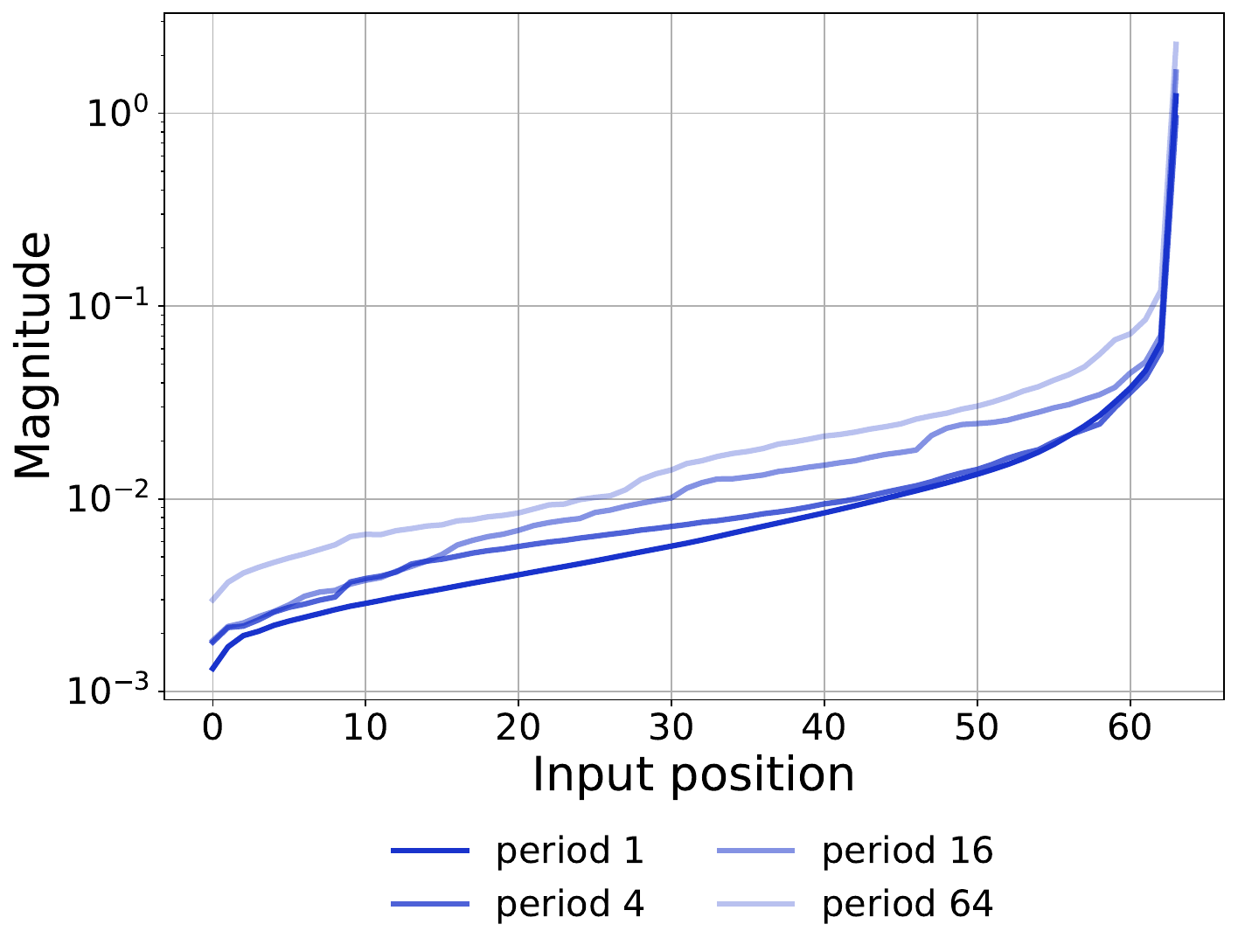}
  \caption{Kernel magnitude across positions under different input periodicities. Lower-frequency (longer-period) inputs yield stronger kernel responses, especially for earlier positions, indicating greater past-time influence.}
  \label{fig:kernels}
\end{figure}

\paragraph{Contributions to Final Token.}
Finally, we compute the kernel (Equation~\ref{eqn:contribution}) showing the contribution of each input position to the last token’s hidden state (Figure~\ref{fig:kernels}). We find that:
\begin{itemize}
    \item Random sequences (\( k=64 \)) show strong contributions from early positions.
    \item Periodic sequences with small \( k \) exhibit weaker contributions.
\end{itemize}

\paragraph{Interpretation.}
Together, these results reveal that Mamba's delta gate adapts to input frequency in a non-trivial way:
\begin{itemize}
    \item \textbf{High-frequency input} \( \Rightarrow \) small delta, slow forgetting, weak input injection.
    \item \textbf{Low-frequency input} \( \Rightarrow \) large delta, fast forgetting, strong input injection.
\end{itemize}
This could explain why repeated relations in the recall task lead to stronger lost-in-middle: the model shifts from attending to inputs to relying more on history.

%% file: sections/conclusions.tex
\section{Discussions}
\label{sec:implications}
Our observation of a U-shaped recall performance in Mamba raises important considerations for the broader use of SSMs in language modeling. While the serial recall task is synthetic and limited in scope, the emergence of such biases may have broader implications, especially for tasks requiring retrieval from arbitrary positions in the context.

First, it is crucial to assess whether this positional bias meaningfully affects downstream performance. For many language tasks, especially those relying on in-context learning, exact copying is not always required. Instead, the model needs to read patterns from the context and generate coherent responses. Our findings show that Mamba still maintains non-zero recall accuracy even in mid-sequence positions, suggesting it can process the full context to some extent. However, for tasks where precise retrieval is critical, such as structured extraction or code generation, this bias could lead to systematic errors.

Second, our results point to a potential architectural limitation in how memory and input are managed. In current Mamba implementations, the same gating factor \( \boldsymbol{\Delta}_t \)  modulates both the recurrent state update (\( \boldsymbol{A}_t \) ) and the input injection (\( \boldsymbol{B}_t \) ). This coupling may constrain the model’s flexibility, especially since (\( \boldsymbol{B}_t \) ) is already input-dependent. We propose that relaxing this coupling (either by decoupling (\( \boldsymbol{B}_t \) ) from \( \boldsymbol{\Delta}_t \)  or introducing an additional gate) could improve memory dynamics. 
\section{Conclusion}
\label{sec:conclusions}



We presented a causal and mechanistic analysis of memory in state-space language models, using primacy and recency as behavioral probes. Our study revealed that Mamba exhibits a robust U-shaped recall pattern, supported by sparse long-term memory channels, fragile short-term dynamics, and input-sensitive gating behavior. These findings challenge the notion that attention is necessary for structured memory and open a new direction for analyzing and improving memory in attention-free models. We hope that this work serves as a foundation for future studies on memory dynamics in SSMs and guides architectural innovations that improve temporal generalization and retrieval fidelity in long-context language modeling.

%% file: sections/appendix.tex
\newpage
\section*{Appendix}
\section{Recall Task Construction and Experimental Setup}
\label{sec:Dataset Contstruction}

\paragraph{Task Construction:}
To evaluate memory behavior in Mamba-based models, we designed a structured recall task consisting of subject-relation-object triplets. Each input sequence contains $L$ unique triplets followed by a single query in the same format (e.g., ``Mike likes''), prompting the model to retrieve the correct object. The queries always target a specific position within the context (e.g., the first or last triplet), allowing us to probe primacy and recency effects.

We ensured that both subject and object tokens were proper names, while relation tokens were verbs commonly used in natural language. Across tasks, we varied the type of relation to prevent overfitting to a specific pattern. Examples include:
\begin{itemize}
    \item \textbf{likes:} Mike likes Todd. Henry likes Victor. $\ldots$ Mike likes, \textbf{answer:}\textit{Todd}
    \item \textbf{Fights:} Austin fights Paula. Jake fights Noah. $\ldots$ Austin fights,\textbf{answer:}\textit{Paula}
    \item \textbf{Affects:} Brian affects Isaac. Amy affects Ron. $\ldots$ Brian affects,\textbf{answer:}\textit{Isaac}
\end{itemize}

Each triplet sequence is tokenized and passed to the model, with accuracy evaluated based on the model's top-1 prediction for the masked query position.

\paragraph{Model Specifications:}
We evaluate two SSM-based models:
\begin{itemize}
    \item \textbf{Mamba 1.4B:} The original open-weight Mamba model, consisting of 48 layers with state-space recurrence \cite{gu2023mamba}.
    \item \textbf{Falcon Mamba 7B:} A larger and more scalable variant adapted from Falcon weights with Mamba-based layers \citep{zuo2024falconmambacompetitiveattentionfree}.
\end{itemize}

Both models are evaluated in inference mode using greedy decoding. For analysis, all subject, relation, and object tokens are constrained to be single-token units. Recall is measured by top-1 accuracy of the predicted object. Although sampling is non-deterministic, we generate 50 examples per target position to ensure robustness and mitigate randomness and lexical artifacts through averaging.

\paragraph{Hardware Setup:}
All experiments were run on a single NVIDIA RTX A6000 GPU with 48GB of memory.

\section{Algorithm for Identifying Long-Term Memory Channels}
We identify long-term memory (LTM) channels in Mamba by measuring their ability to preserve early input information through recurrent dynamics. The core hypothesis is that channels exhibiting strong cumulative recurrence from the first timestep act as long-term memory.

Let \( A_t^{(i)} \in \mathbb{R}^{N \times N} \) and \( B_t^{(i)} \in \mathbb{R}^{N} \) denote the recurrence matrix and input projection at timestep \( t \) for channel \( i \), respectively. For each channel, we compute the cumulative product of \( A_t^{(i)} \) across time and determine the proportion of dimensions exceeding a recurrence threshold \( \tau \). A channel is marked as an LTM candidate if this proportion exceeds a threshold \( p \).

Algorithm~\ref{alg:contribution-filter} summarizes the procedure. A forward pass on a single input sample is used to extract the relevant matrices. The output is a dictionary \texttt{layers\_dim} mapping each layer to its LTM-identified channels, which are later used for targeted intervention (Section~\ref{sec:primacy}).

\begin{algorithm}[H]
\caption{Identifying High-Contribution Dimensions Across Layers}
\label{alg:contribution-filter}
\begin{algorithmic}
\Require Input sequence, recurrence matrices \( A_t \), threshold \( \tau \), proportion \( p \)
\State Initialize \texttt{layers\_dim} as a dictionary for storing selected channels per layer
\State\textbf{do} Perform a forward pass to collect \( A_t \) and \( B_t \) across all timesteps
\For{each layer \( \ell = 0 \) to \( L{-}1 \)}
    \State \texttt{satisfying\_dim} \( \gets [\,] \)
    \For{each channel \( d = 0 \) to \( D{-}1 \)}
        \State \( A \gets A_\ell[:, d, :t, :] \) \Comment{Recurrence across time}
        \State \( M \gets \prod_{t=2}^{T} A_t \) \Comment{Element-wise product across time}
        \State \( \rho \gets \text{mean}(\mathbb{I}[M > \tau]) \)
        \If{\( \rho > p \)}
            \State Append \( (d, \rho) \) to \texttt{satisfying\_dim}
        \EndIf
    \EndFor
    \State \texttt{layers\_dim[\( \ell \)]} \( \gets \) \texttt{satisfying\_dim}
\EndFor
\end{algorithmic}
\end{algorithm}

\section{Ablation of the choice of $p$ and $\tau$}
We find that high-precision selection (\( p = 0.9 \)) with a lenient threshold (\( \tau = 0.5 \)) yields the greatest drop in recall. Increasing the number of intervened layers further amplifies this effect.

\label{sec: tau_p_ablation}
\begin{figure}[H]
  \includegraphics[width=\columnwidth]{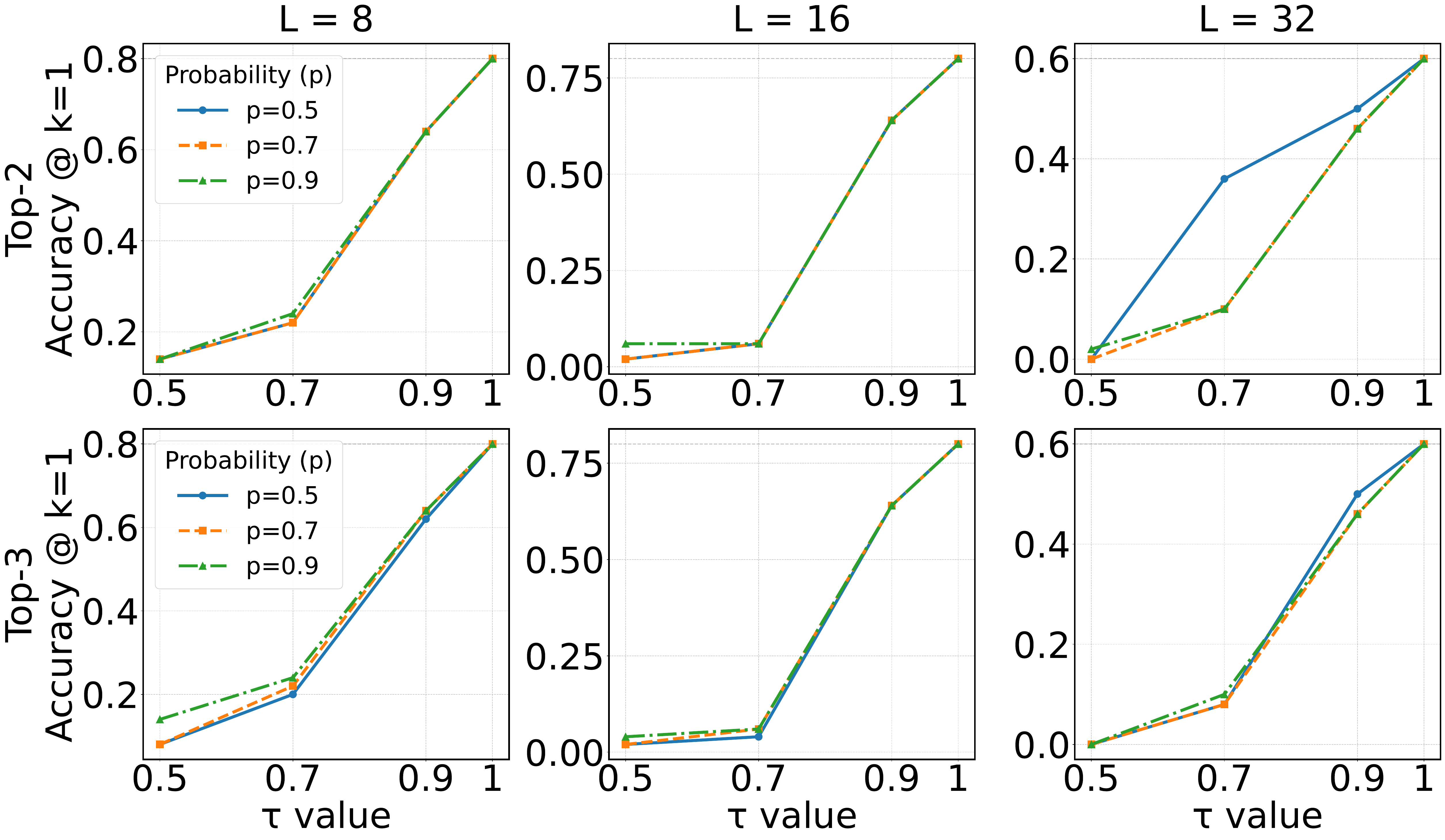}
  \caption{Ablation of the choice of $p$, $\tau$ at intervening at top-n layers}
  \label{fig:intervention16}
\end{figure}

\section{Intervention on Longer Sequences}
\label{sec:Intervention_16_32}

\begin{figure}[H]
  \centering
  \begin{subfigure}{\columnwidth}
    \centering
    \includegraphics[width=\textwidth]{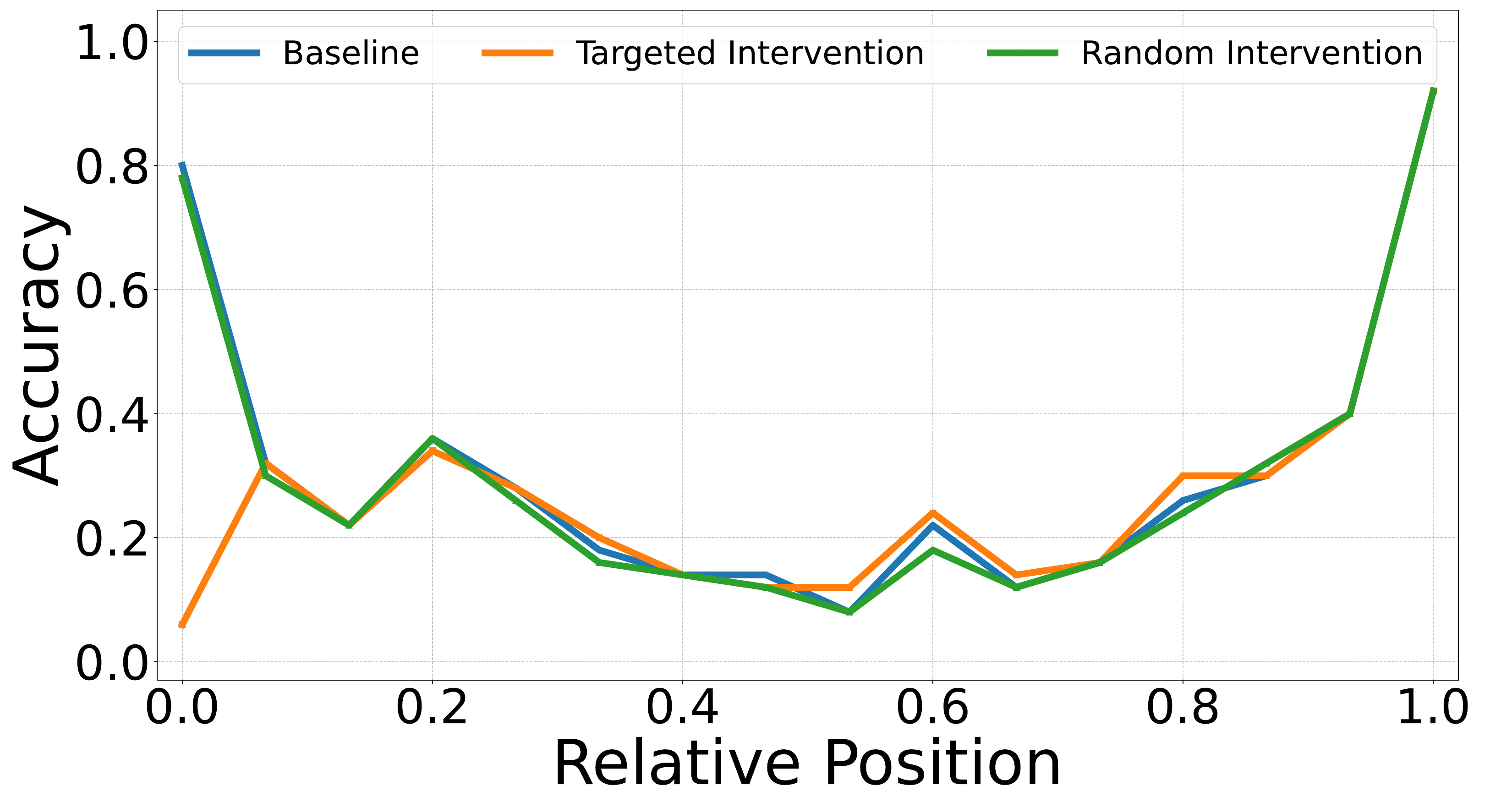}
    \caption{Intervention on L=16}
    \label{fig:intervention16}
  \end{subfigure}
  
  \vspace{1em}
  
  \begin{subfigure}{\columnwidth}
    \centering
    \includegraphics[width=\textwidth]{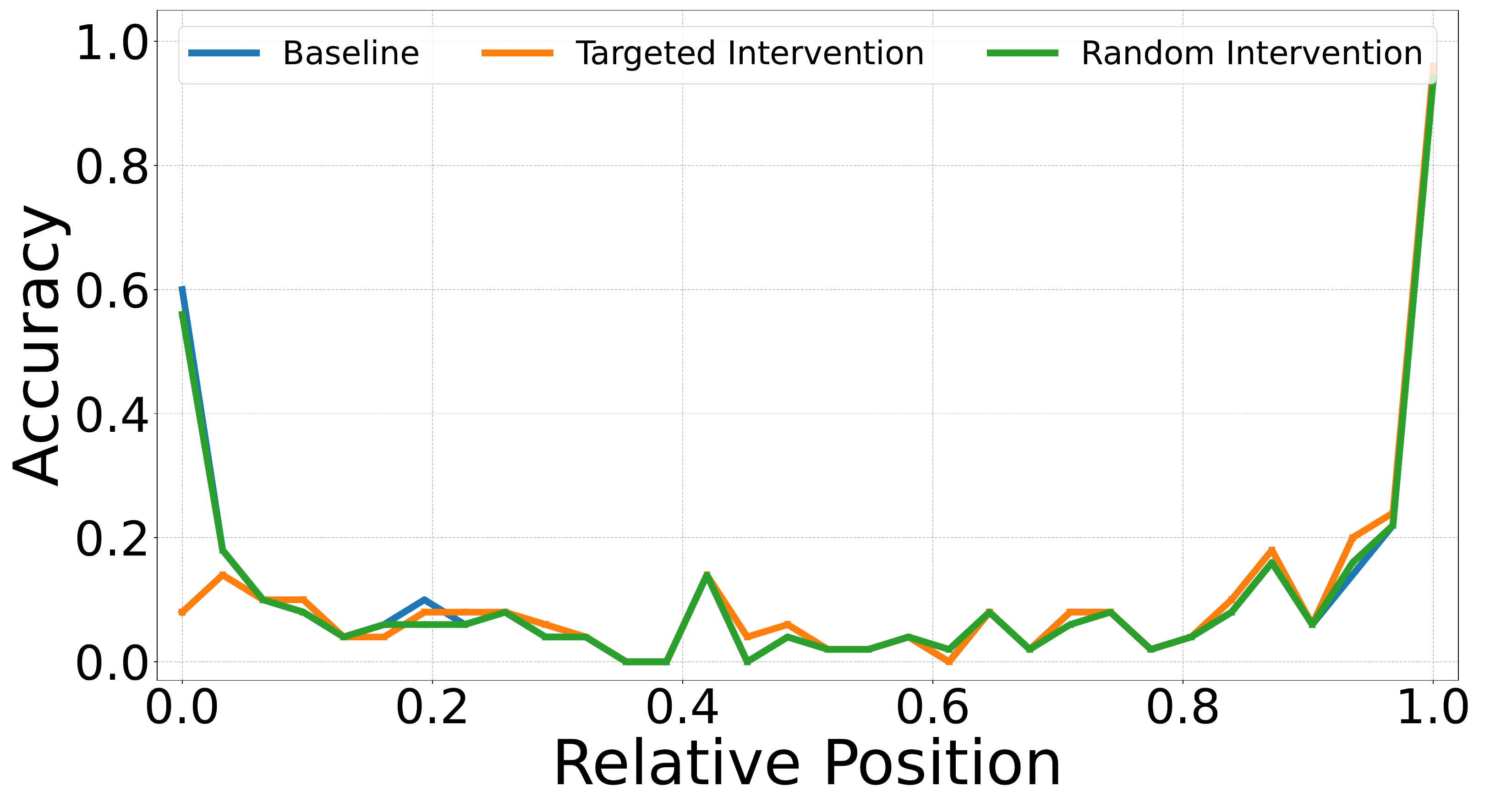}
    \caption{Intervention on L=32}
    \label{fig:intervention32}
  \end{subfigure}
  \caption{Intervention results for different length of sequence.}
  \label{fig:intervention_16_32}
\end{figure}

We also evaluate the intervention method on longer sequences (\( L = 16 \), \( L = 32 \)) using the same parameters (\( p = 0.7 \), \( \tau = 0.7 \)). As shown in Figure~\ref{fig:intervention_16_32}, the intervention remains effective: recall at the first position drops substantially, indicating that the identified long-term memory channels continue to play a critical role as sequence length increases.

\section{Results on Different Model Size: Mamba 1.4B}
To test the generalization of our proposed method of localization and intervention, we conducted the same experiments in a different model variant, namely Mamba 1.4 B.

\subsection{Ablation of the choice of $p$ and $\tau$}
We begin by examining the behavior of different \( p \) and \( \tau \) configurations in Mamba 1.4B. The patterns observed in Falcon Mamba 7B hold consistently: even for shorter sequences, high-precision selection (e.g., \( p = 0.9 \), \( \tau = 0.7 \)) already results in a noticeable drop in early recall. However, to achieve stronger intervention effects in Mamba 1.4B, it becomes necessary to increase the number of layers being ablated. This is likely due to the more distributed nature of long-term memory channels in smaller models. With fewer total parameters, the model may be forced to allocate long-term memory more sparsely across layers, requiring broader intervention to disrupt its recall capacity effectively.

\begin{figure}[H]
  \includegraphics[width=\columnwidth]{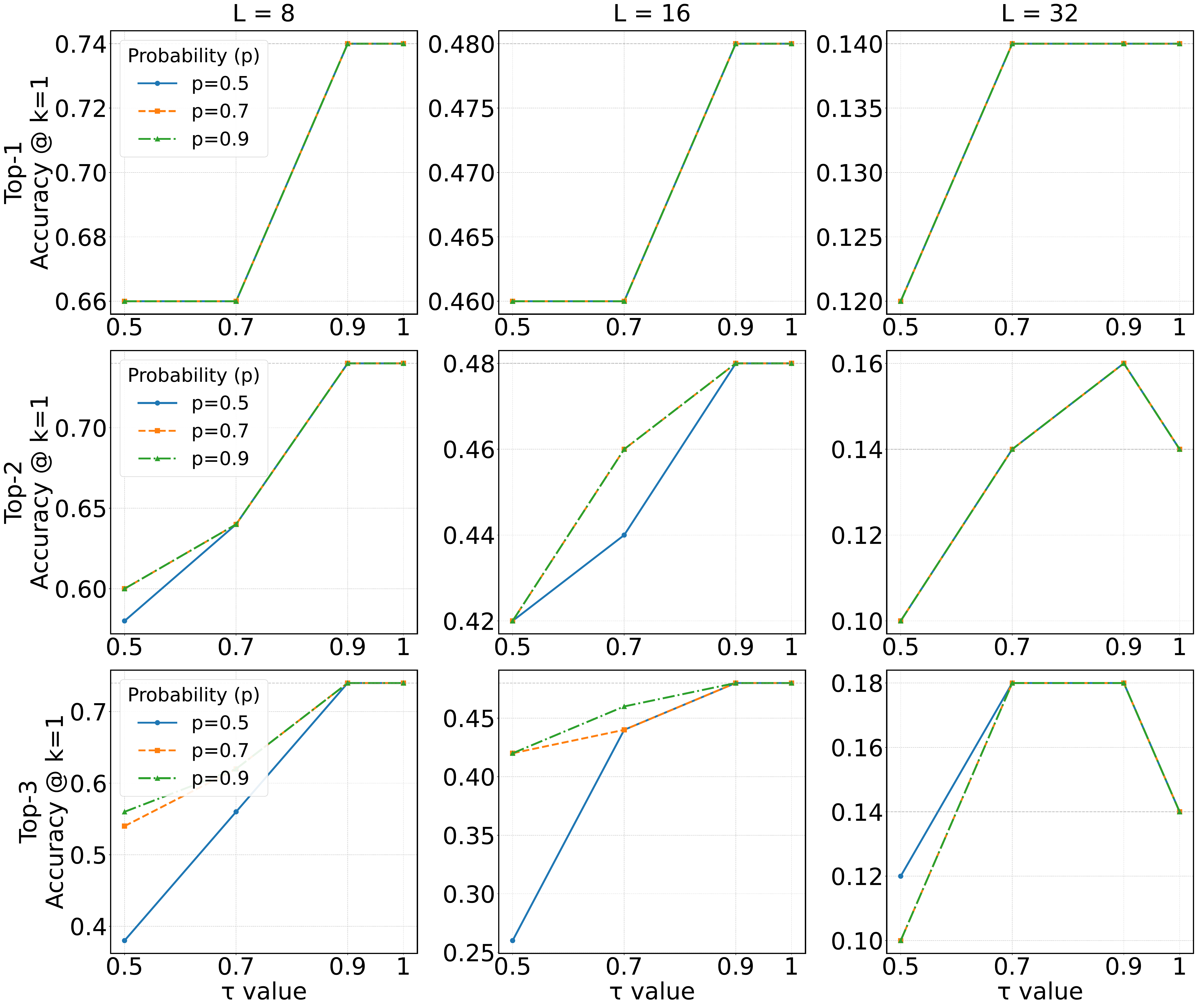}
  \caption{Ablation of the choice of $p$, $\tau$ at intervening at top-n layers for Mamba 1.4B}
  \label{fig:intervention16}
\end{figure}
\subsection{Intervention Result on Various Length}
\begin{figure}[H]
  \centering
  \begin{subfigure}{\columnwidth}
    \centering
    \includegraphics[width=\textwidth]{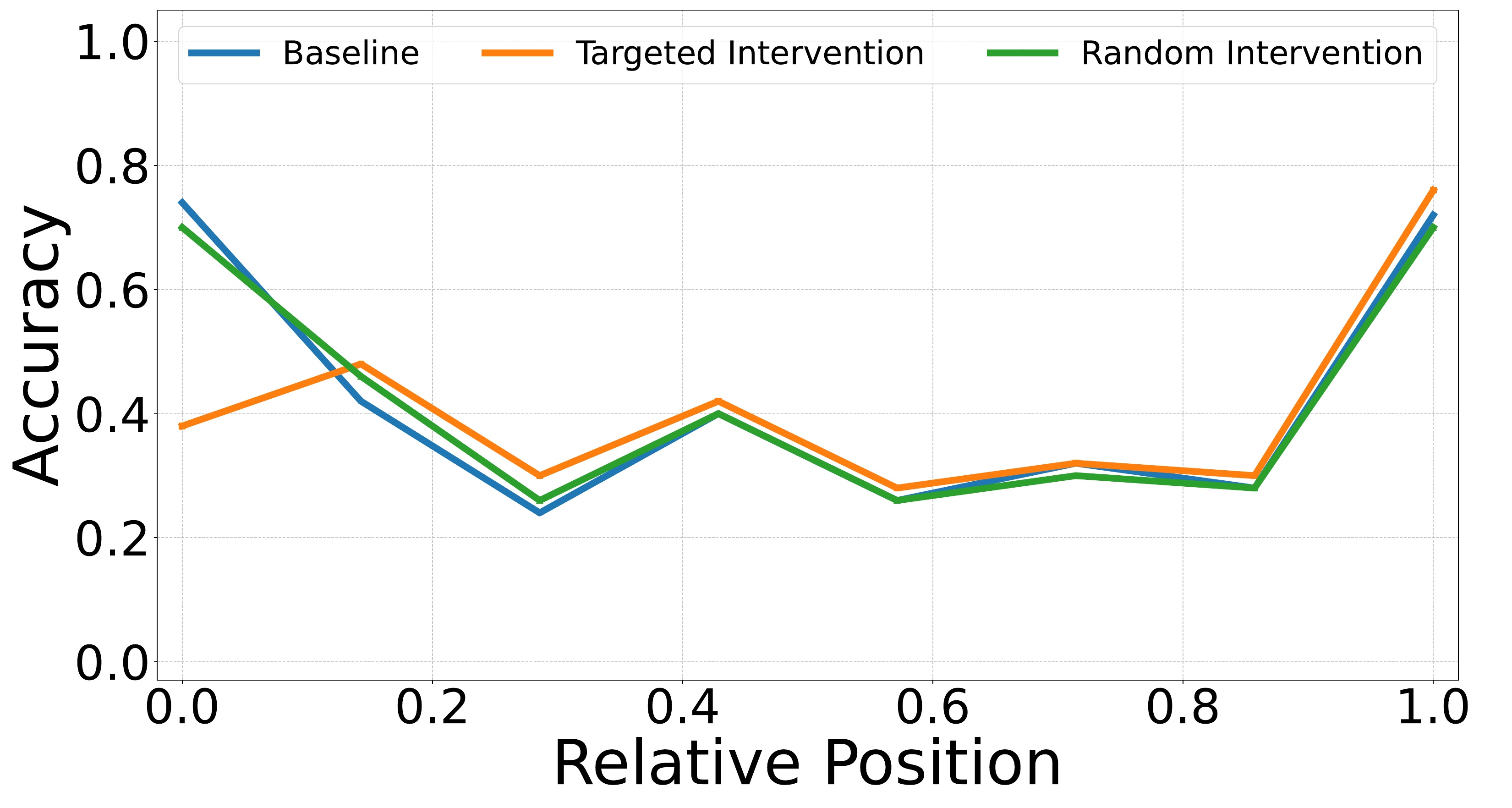}
    \caption{Intervention on L=8}
    \label{fig:mamba1b_intervention8}
  \end{subfigure}
  
  \vspace{1em}
  
  \begin{subfigure}{\columnwidth}
    \centering
    \includegraphics[width=\textwidth]{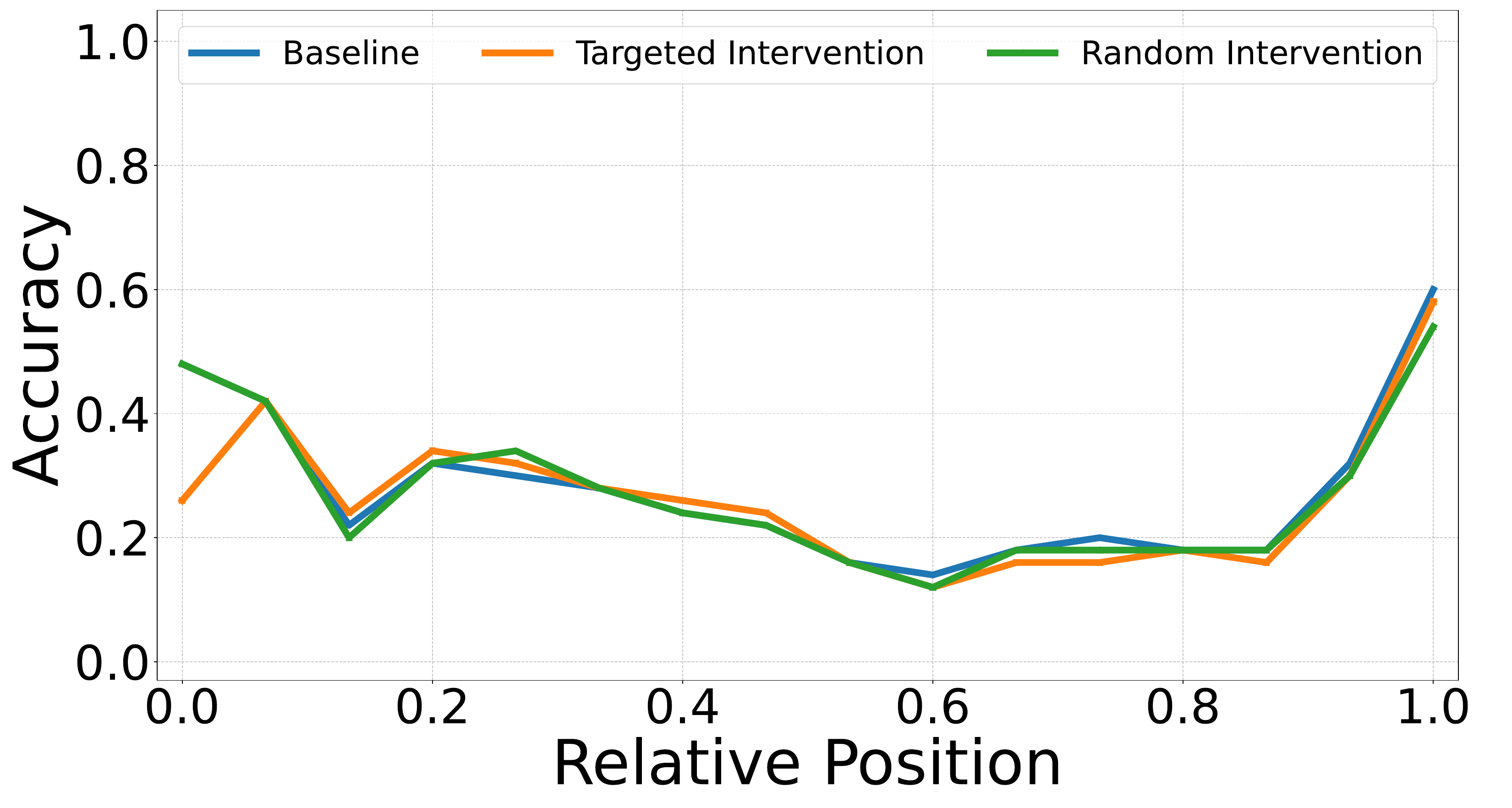}
    \caption{Intervention on L=16}
    \label{fig:mamba1b_intervention16}
  \end{subfigure}
  
  \vspace{1em}
  
  \begin{subfigure}{\columnwidth}
    \centering
    \includegraphics[width=\textwidth]{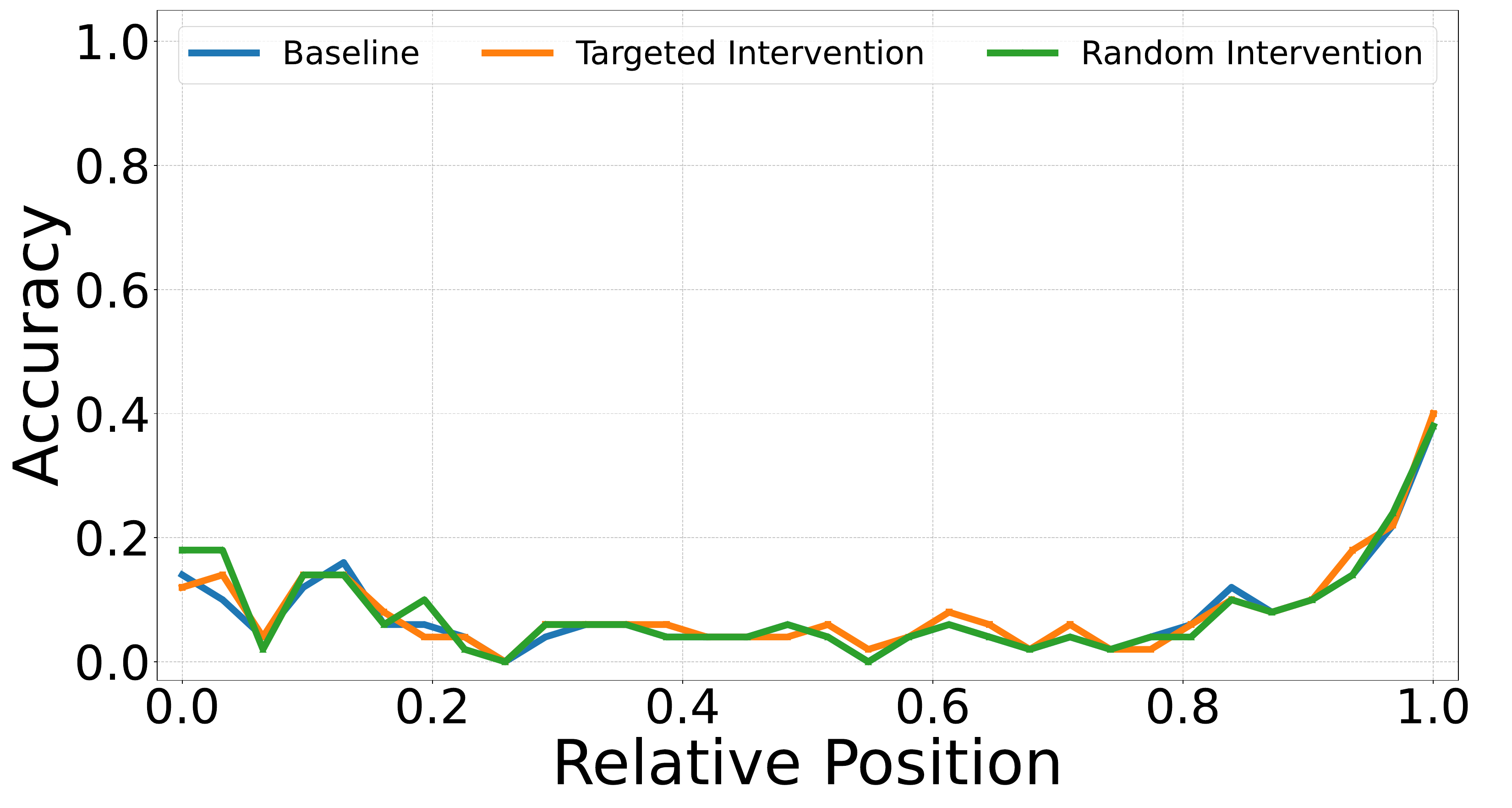}
    \caption{Intervention on L=32}
    \label{fig:mamba1b_intervention32}
  \end{subfigure}
  \caption{Intervention results for different lengths of sequence on Mamba 1.4B}
  \label{fig:mamba1b_intervention_combined}
\end{figure}

We also apply the proposed intervention method to Mamba 1.4B. Based on the earlier ablation results, the largest drop in accuracy at \( k = 1 \) was observed using parameters \( p = 0.5 \), \( \tau = 0.5 \), and intervention on the top-3 layers. As shown in Figure~\ref{fig:mamba1b_intervention_combined}, the intervention effectively impairs the model’s ability to recall the object from the first triplet, demonstrating the generalizability of our identification method. However, for longer sequences (e.g., \( L = 32 \)), the drop is less pronounced, likely because the model already struggles to retain early input in such settings—even without intervention.

%% file: acl_latex.bbl
\begin{thebibliography}{16}
\providecommand{\natexlab}[1]{#1}

\bibitem[{Atkinson and Shiffrin(1968)}]{atkinson1968human}
Richard~C Atkinson and Richard~M Shiffrin. 1968.
\newblock Human memory: A proposed system and its control processes.
\newblock In \emph{Psychology of learning and motivation}, volume~2, pages 89--195. Elsevier.

\bibitem[{Ebbinghaus(1913)}]{ebbinghaus1885memory}
Hermann Ebbinghaus. 1913.
\newblock Memory: A contribution to experimental psychology (1913).
\newblock \emph{New York: Teachers College, Columbia University}.

\bibitem[{Glanzer and Cunitz(1966)}]{glanzer1966two}
Murray Glanzer and Anita~R Cunitz. 1966.
\newblock Two storage mechanisms in free recall.
\newblock \emph{Journal of verbal learning and verbal behavior}, 5(4):351--360.

\bibitem[{Gu and Dao(2023)}]{gu2023mamba}
Albert Gu and Tri Dao. 2023.
\newblock Mamba: Linear-time sequence modeling with selective state spaces.
\newblock \emph{arXiv preprint arXiv:2312.00752}.

\bibitem[{Gu et~al.(2022)Gu, Goel, Gupta, and R{\'e}}]{gu2022parameterization}
Albert Gu, Karan Goel, Ankit Gupta, and Christopher R{\'e}. 2022.
\newblock On the parameterization and initialization of diagonal state space models.
\newblock \emph{Advances in Neural Information Processing Systems}, 35:35971--35983.

\bibitem[{Gu et~al.(2021)Gu, Goel, and R{\'e}}]{gu2021efficiently}
Albert Gu, Karan Goel, and Christopher R{\'e}. 2021.
\newblock Efficiently modeling long sequences with structured state spaces.
\newblock \emph{arXiv preprint arXiv:2111.00396}.

\bibitem[{Gu et~al.(2025)Gu, Pang, Du, Liu, Zhang, Du, Wang, and Lin}]{gu2025attentionsinkemergeslanguage}
Xiangming Gu, Tianyu Pang, Chao Du, Qian Liu, Fengzhuo Zhang, Cunxiao Du, Ye~Wang, and Min Lin. 2025.
\newblock \href {https://arxiv.org/abs/2410.10781} {When attention sink emerges in language models: An empirical view}.
\newblock \emph{Preprint}, arXiv:2410.10781.

\bibitem[{Jahnke(1969)}]{jahnke1969ranschburg}
John~C Jahnke. 1969.
\newblock The ranschburg effect.
\newblock \emph{Psychological Review}, 76(6):592.

\bibitem[{Janik(2023)}]{janik2023aspects}
Romuald~A Janik. 2023.
\newblock Aspects of human memory and large language models.
\newblock \emph{arXiv preprint arXiv:2311.03839}.

\bibitem[{Jelassi et~al.(2024)Jelassi, Brandfonbrener, Kakade, and Malach}]{jelassi2024repeat}
Samy Jelassi, David Brandfonbrener, Sham~M Kakade, and Eran Malach. 2024.
\newblock Repeat after me: Transformers are better than state space models at copying.
\newblock \emph{arXiv preprint arXiv:2402.01032}.

\bibitem[{Liu and Li(2024)}]{liu2024autocorrelation}
Fusheng Liu and Qianxiao Li. 2024.
\newblock Autocorrelation matters: Understanding the role of initialization schemes for state space models.
\newblock \emph{arXiv preprint arXiv:2411.19455}.

\bibitem[{Morita(2025)}]{morita2025emergenceprimacyeffectstructured}
Takashi Morita. 2025.
\newblock \href {https://arxiv.org/abs/2502.13729} {Emergence of the primacy effect in structured state-space models}.
\newblock \emph{Preprint}, arXiv:2502.13729.

\bibitem[{Olsson et~al.(2022)Olsson, Elhage, Nanda, Joseph, DasSarma, Henighan, Mann, Askell, Bai, Chen et~al.}]{olsson2022context}
Catherine Olsson, Nelson Elhage, Neel Nanda, Nicholas Joseph, Nova DasSarma, Tom Henighan, Ben Mann, Amanda Askell, Yuntao Bai, Anna Chen, and 1 others. 2022.
\newblock In-context learning and induction heads.
\newblock \emph{arXiv preprint arXiv:2209.11895}.

\bibitem[{Wang et~al.(2025)Wang, Cai, Wang, Zhu, Srivastava, Wang, and Li}]{wang2025understandingmitigatingbottlenecksstate}
Peihao Wang, Ruisi Cai, Yuehao Wang, Jiajun Zhu, Pragya Srivastava, Zhangyang Wang, and Pan Li. 2025.
\newblock \href {https://arxiv.org/abs/2501.00658} {Understanding and mitigating bottlenecks of state space models through the lens of recency and over-smoothing}.
\newblock \emph{Preprint}, arXiv:2501.00658.

\bibitem[{Wu et~al.(2025)Wu, Wang, Jegelka, and Jadbabaie}]{wu2025emergence}
Xinyi Wu, Yifei Wang, Stefanie Jegelka, and Ali Jadbabaie. 2025.
\newblock On the emergence of position bias in transformers.
\newblock \emph{arXiv preprint arXiv:2502.01951}.

\bibitem[{Zuo et~al.(2024)Zuo, Velikanov, Rhaiem, Chahed, Belkada, Kunsch, and Hacid}]{zuo2024falconmambacompetitiveattentionfree}
Jingwei Zuo, Maksim Velikanov, Dhia~Eddine Rhaiem, Ilyas Chahed, Younes Belkada, Guillaume Kunsch, and Hakim Hacid. 2024.
\newblock \href {https://arxiv.org/abs/2410.05355} {Falcon mamba: The first competitive attention-free 7b language model}.
\newblock \emph{Preprint}, arXiv:2410.05355.

\end{thebibliography}
